\pdfoutput=1
\PassOptionsToPackage{table,dvipsnames}{xcolor}

\documentclass[11pt]{article}

\usepackage[final]{acl}

\usepackage{times}
\usepackage{latexsym}

\usepackage[T1]{fontenc}

\usepackage[utf8]{inputenc}

\usepackage{microtype}

\usepackage{inconsolata}

\usepackage{graphicx}
\usepackage{caption}
\usepackage{subcaption}
\usepackage{pifont}
\newcommand{\cmark}{\ding{51}}   
\newcommand{\xmark}{\ding{55}}  

\usepackage{booktabs} 

\usepackage{xspace}
\usepackage[most]{tcolorbox}

\usepackage{enumitem}  
\usepackage{pifont}
\usepackage{amsmath}
\usepackage{amssymb} 

\usepackage{graphicx}   
\usepackage{booktabs}   
\usepackage{multirow}   
\usepackage{arydshln}   
\usepackage{array}
\usepackage{tabularx}
\usepackage{adjustbox} 
\usepackage{xcolor}

\definecolor{green2}{RGB}{90, 130, 60}
\definecolor{gray2}{RGB}{169, 169, 169}
\definecolor{red2}{RGB}{178, 34, 34}
\definecolor{mygreen2}{RGB}{238, 243, 243}

\usepackage{listings}
\usepackage{bm}
\usepackage{tcolorbox}
\newcommand{\dataset}{\textsc{PhysicsArena}\xspace}
\tcbuselibrary{listings}

\lstdefinelanguage{json}{
  basicstyle   = \footnotesize\ttfamily,
  showstringspaces = false,
  breaklines   = true,
  morecomment  = [l]{//},
  commentstyle = \color{gray}\itshape,
  stringstyle  = \color{teal},
  morestring   = [s]{"}{"},
}

\title{\dataset: The First Multimodal Physics Reasoning Benchmark Exploring Variable, Process, and Solution Dimensions}

\author{Song Dai\textsuperscript{\rm 1,\rm 2,*}, Yibo Yan\textsuperscript{\rm 1,\rm 3,*}, Jiamin Su\textsuperscript{\rm 1,\rm 2}, Zihao Dongfang\textsuperscript{\rm 1}, 
\textbf{Yubo Gao}\textsuperscript{\rm 1}, 
\textbf{Yonghua Hei}\textsuperscript{\rm 1,\rm 2,\rm 3}, \\
\textbf{Jungang Li}\textsuperscript{\rm 1,\rm 2},
\textbf{Junyan Zhang}\textsuperscript{\rm 1},
\textbf{Sicheng Tao}\textsuperscript{\rm 1},
\textbf{Zhuoran Gao}\textsuperscript{\rm 1,\rm 2,\rm 3},
\textbf{Xuming Hu}\textsuperscript{\rm 1,\rm 2,\rm 3,}\footnotemark[2]\\
\textsuperscript{\rm 1}The Hong Kong University of Science and Technology (Guangzhou)\\
\textsuperscript{\rm 2}Beijing Future Brain Education Technology Co., Ltd.\\
\textsuperscript{\rm 3}The Hong Kong University of Science and Technology\\
\texttt{\href{mailto:samdie2016@gmail.com}{\{samdie2016}, \href{mailto:yanyibo70@gmail.com}{yanyibo70\}@gmail.com}}, 
     \texttt{\href{mailto:xuminghu@hkust-gz.edu.cn}{xuminghu@hkust-gz.edu.cn}}
}

\begin{document}
\maketitle
\renewcommand{\thefootnote}{\fnsymbol{footnote}}
\footnotetext[1]{Co-first authors with equal contribution.}
\footnotetext[2]{Corresponding author.}
\renewcommand{\thefootnote}{\arabic{footnote}}
\begin{abstract}
Multimodal Large Language Models (MLLMs) have demonstrated remarkable capabilities in diverse reasoning tasks, yet their application to complex physics reasoning remains underexplored.
Physics reasoning presents unique challenges, requiring grounding in physical conditions and the interpretation of multimodal information.
Current physics benchmarks are limited, often focusing on \textit{text-only inputs} or \textit{solely on problem-solving}, thereby overlooking the critical intermediate steps of variable identification and process formulation.
To address these limitations, we introduce \dataset, the first multimodal physics reasoning benchmark designed to holistically evaluate MLLMs across three critical dimensions: \textit{variable identification}, \textit{physical process formulation}, and \textit{solution derivation}.
PhysicsArena aims to provide a comprehensive platform for assessing and advancing the multimodal physics reasoning abilities of MLLMs.
\end{abstract}

\section{Introduction}

Multimodal Large Language Models (MLLMs) have recently demonstrated remarkable capabilities across a diverse range of domains~\cite{caffagni2024revolution,fei2024multimodal,yan2024urbanclip,yan2024errorradar}. Their proficiency in processing and integrating information from various modalities has unlocked significant potential~\cite{fu2024mme,huo2024mmneuron}. Notably, the reasoning abilities inherent in the underlying LLMs have fueled advancements in multimodal reasoning tasks. This synergy is particularly beneficial in complex, real-world scenarios such as education, where understanding and reasoning about multimodal information are paramount. Areas like mathematical problem-solving and code generation have already seen substantial progress, showcasing the power of MLLMs in tackling structured reasoning challenges~\cite{yan2024survey,yun2024web2code,wang2024exploring,lin2025mind}.

\begin{figure}[t!]
  \centering
  \includegraphics[width=0.48\textwidth]{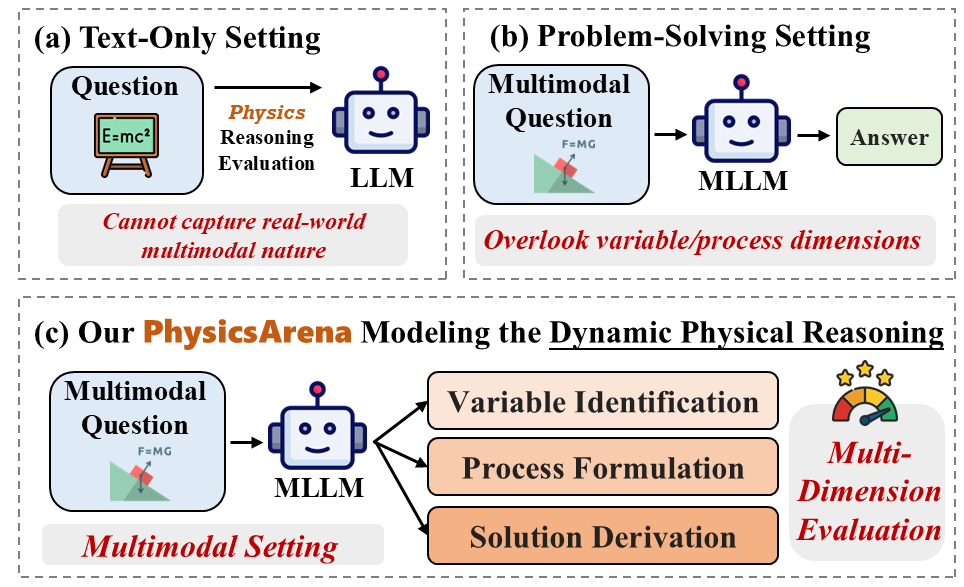}
  \caption{Comparison between previous physics reasoning settings and our proposed \dataset.}
  \label{fig:paradigm_comparison}
  \vspace{-4mm}
\end{figure}

Despite these advancements, the \textit{domain of physics reasoning remains relatively underexplored within the MLLM research landscape}. Physics presents a unique and arguably more intricate reasoning setting compared to mathematics or coding. Effective physics reasoning necessitates not only logical deduction but also a deep understanding \textbf{grounded in real-world physical laws and theorems}. Furthermore, the reasoning process is often tightly \textbf{constrained by objective physical conditions} depicted visually or described textually. This inherent complexity, involving the interplay between abstract principles and concrete, often multimodal, scenarios, necessitates a dedicated benchmark capable of rigorously evaluating the physics reasoning capabilities of modern MLLMs~\cite{yan2025position,ferrag2025reasoning}.

Current benchmarks designed for physics reasoning suffer from significant limitations as follows. \ding{182} Many existing efforts~\cite{qiuPHYBenchHolisticEvaluation2025,xuUGPhysicsComprehensiveBenchmark2025} primarily focus on text-only settings, \textit{failing to capture the crucial interplay with multimodal information that characterizes real-world physics problems} (e.g., interpreting diagrams, graphs, or experimental setups), as shown in Figure~\ref{fig:paradigm_comparison} (a). \ding{183} Other benchmarks~\cite{fengPHYSICSBenchmarkingFoundation2025,zhangPhysReasonComprehensiveBenchmark2025}, even if multimodal, tend to \textit{concentrate solely on the problem-solving aspect} – predicting the final solution or answer, as shown in Figure~\ref{fig:paradigm_comparison} (b). This overlooks the critical intermediate steps inherent in physics reasoning: identifying relevant variables from the problem context and formulating the correct physical process or sequence of principles required to reach the solution. A comprehensive evaluation of physics reasoning capabilities, therefore, requires modeling the dynamic reasoning process from its inception, encompassing these vital variable and process stages.

\begin{table*}[t!]
  \small
  \setlength{\tabcolsep}{4pt}
  \centering
  
  \begin{tabular}{lrrllccc}
    \toprule
    \textbf{Benchmarks} & \textbf{Size} & \textbf{Img.\#} &
    \textbf{Know. Level} & \textbf{Qns. Type} &
    \multicolumn{3}{c}{\textbf{Task Dimension}}\\
    & & & & & \textbf{Variable} & \textbf{Process} & \textbf{Solution}\\
    \midrule  
    E-EVAL ~\cite{houEEVALComprehensiveChinese2024}        &   342 &     0 & K12   & MC     & \xmark & \xmark & -- \\
    MMLU-Pro~\cite{wangMMLUProMoreRobust2024}     & 1299 &     0 & COL      & MC     & \xmark & \xmark & -- \\
    GPQA~\cite{reinGPQAGraduateLevelGoogleProof2024}         &   227 &     0 & Ph.D     & OE     & \xmark & \xmark & \cmark \\
    JEEBench~\cite{aroraHaveLLMsAdvanced2023a}     &   123 &     0 & CEE      & OE/MC  & \xmark & \xmark & \cmark \\
    ScienceQA~\cite{luLearnExplainMultimodal2022}    & 1923 & 1328 & K12   & MC     & \xmark & \xmark & -- \\
    SciBench~\cite{wangSciBenchEvaluatingCollegeLevel2023}     &   291 &    64 & COL      & OE     & \xmark & \xmark & \cmark \\
    MMMU~\cite{yueMMMUMassiveMultiDiscipline2024}         &   443 &   443 & COL      & OE/MC  & \xmark & \xmark & \cmark \\
    OlympiadBench~\cite{heOlympiadBenchChallengingBenchmark2024} & 2334 & 1958 & CEE/COMP & OE     & \xmark & \xmark & \cmark \\
    SciEval~\cite{sunSciEvalMultiLevelLarge2024}      & 1657 &     0 & --       & OE/MC  & \xmark & \xmark & \cmark \\
    EMMA~\cite{hao2025mllmsreasonmultimodalityemma}         &   156 &   156 & CEE      & MC     & \xmark & \xmark & -- \\
    PhyReason~\cite{zhangPhysReasonComprehensiveBenchmark2025}    & 1200 &   972 & CEE/COMP & OE     & \xmark & \xmark & \cmark \\
    PHYBench~\cite{qiuPHYBenchHolisticEvaluation2025}     &   500 &     0 & COMP/COL & OE     & \xmark & \xmark & \cmark \\
    PHYSICS~\cite{fengPHYSICSBenchmarkingFoundation2025}     & 1297 &   298 & COL      & OE     & \xmark & \xmark & \cmark \\
    UGPhysics~\cite{xuUGPhysicsComprehensiveBenchmark2025}     &  5520&  0  &  UG     & OE/MC     & \xmark & \xmark & \cmark \\
    \midrule
     {\textbf{\dataset (Ours)}} & {5103} & {5103} & {CEE}      & {OE}     & {\cmark} & {\cmark} & {\cmark} \\
    \bottomrule
  \end{tabular}
  \caption{Comparisons between physics reasoning benchmark (covering the physics-related data included in scientific reasoning benchmarks) vs our proposed \dataset dataset.
          \textbf{Img.\#:} Count of problems with image;
          \textbf{Knowledge Level:} \textit{K12}: Elementary to High School; \textit{CEE}: College Entrance Examination;  \textit{COMP}: Competition; \textit{COL}: College; \textit{UG}: Undergraduate; \textit{Ph.D}: Doctor of Philosophy.
          \textbf{Question Type:} \textit{OE}: Open-ended; \textit{MC}: Multiple-choice.}
    \label{tab:benchmarks}
\end{table*}
To bridge this gap, we introduce \textbf{\dataset, the first benchmark specifically designed to comprehensively evaluate multimodal physics reasoning} across three crucial dimensions: \textit{Variable Identification}, \textit{Process Formulation}, and \textit{Solution Derivation}. As illustrated in Figure~\ref{fig:paradigm_comparison} (c), \dataset provides a structured environment with problems presented multimodally, demanding that models demonstrate understanding throughout the entire reasoning pipeline, not just at the final output stage. By dissecting the reasoning task into these three interconnected dimensions, our benchmark offers a more granular and insightful assessment of MLLM capabilities in this challenging domain. We have rigorously evaluated a suite of representative, state-of-the-art MLLMs using \dataset.

Our contributions can be summarized as follows:
\begin{itemize}[leftmargin=*]
    \item We introduce \dataset, the first multimodal physics reasoning benchmark that explicitly models the dynamic reasoning process. It comprises over 5,000 high-quality instances.
    \item \dataset provides a holistic evaluation framework by incorporating assessments across the Variable, Process, and Solution dimensions. This multi-faceted approach fully addresses the complexity inherent in the physical setting.
    \item We conduct extensive experiments on representative state-of-the-art MLLMs using \dataset. Our results provide valuable insights into capabilities, revealing a significant gap that still exists towards AGI-level intelligence.
\end{itemize}


\section{Related Works}

\subsection{Physics Reasoning Benchmarks}
As the community's focus on scientific reasoning increases~\cite{luo2025llm4sr,yan2025mathagent,yan2024georeasoner}, physics reasoning also requires high-quality benchmarks for evaluation. As indicated in Table~\ref{tab:benchmarks}, early physics reasoning data were all subsets of general scientific reasoning benchmarks. Early science‑wide suites such as
E‑EVAL~\cite{houEEVALComprehensiveChinese2024} for Chinese K‑12 education, MMLU‑Pro~\cite{wangMMLUProMoreRobust2024} for
college‑level knowledge, and the multimodal ScienceQA dataset~\cite{luLearnExplainMultimodal2022}
establish broad coverage with text‑only or image‑augmented multiple‑choice questions
across diverse subjects that include physics. Subsequent resources raise disciplinary
depth: GPQA~\cite{reinGPQAGraduateLevelGoogleProof2024} introduces graduate‑level STEM questions designed to be
\textit{Google‑proof}; JEEBench~\cite{aroraHaveLLMsAdvanced2023a} curates IIT·JEE‑Advanced problems
combining MC and open‑ended formats; and college‑focused sets such as
SciBench~\cite{wangSciBenchEvaluatingCollegeLevel2023}, SciEval~\cite{sunSciEvalMultiLevelLarge2024}, and the bilingual
multimodal OlympiadBench~\cite{heOlympiadBenchChallengingBenchmark2024} adopt numerical or free‑response
answers and often supply diagram contexts. In the past year, reasoning benchmarks specifically dedicated to physics have begun to emerge.
PhysReason~\cite{zhangPhysReasonComprehensiveBenchmark2025} provides 1,200 problems with step‑level
assessment, PHYBench~\cite{qiuPHYBenchHolisticEvaluation2025} introduces an expression‑distance metric
over 500 real‑world scenarios, and UGPhysics~\cite{xuUGPhysicsComprehensiveBenchmark2025} couples 5,520 undergraduate problems with a rule‑based judgment pipeline. Together these benchmarks trace a coherent evolution from general science to domain‑focused physics, from
fixed‑choice to open‑ended solutions, and from text to richly multimodal settings~\cite{chen2025towards,li2025system}.

\begin{figure*}[t!]
    \centering
    \includegraphics[width=0.75\textwidth]{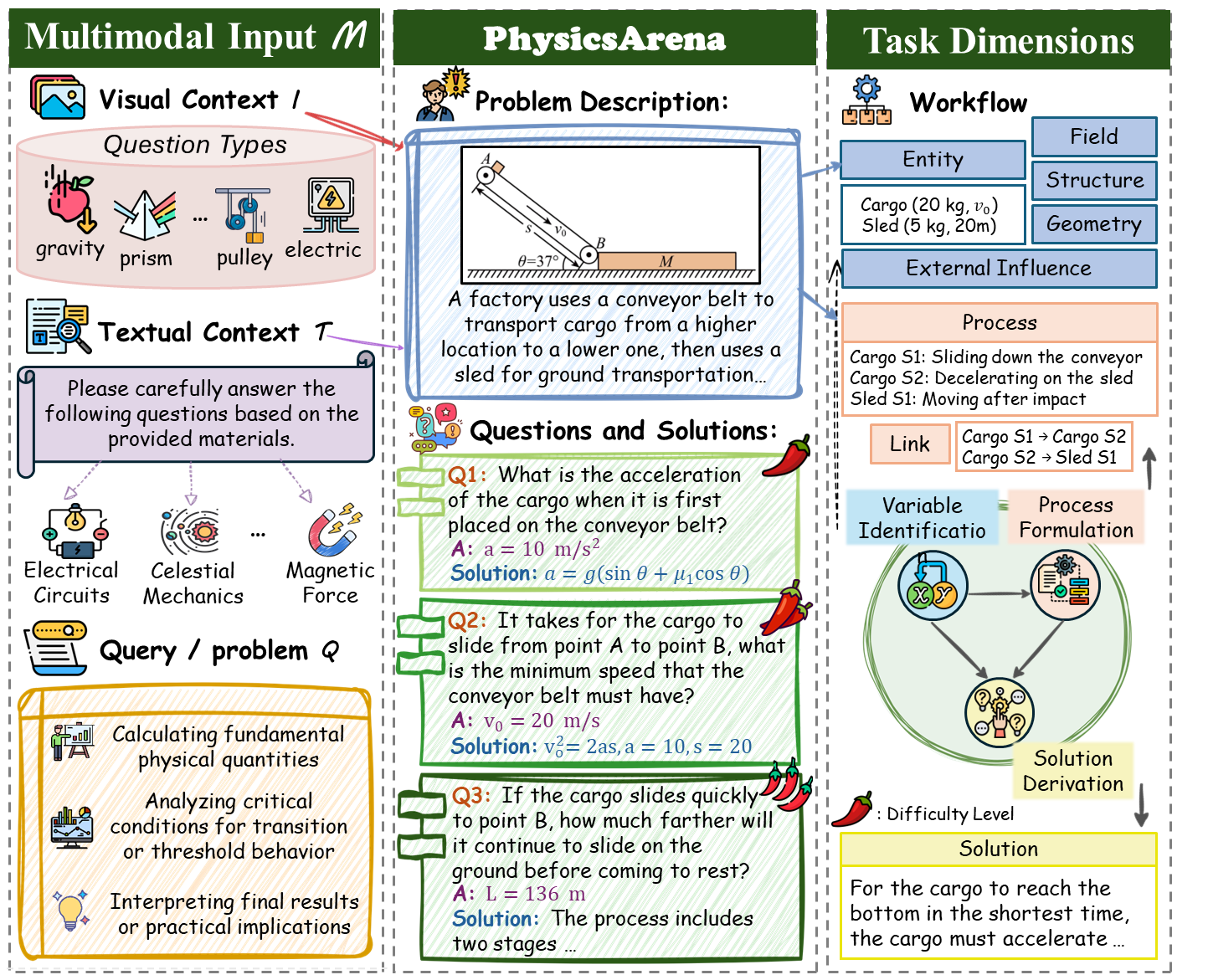}
    \caption{The illustration of a representative example from our proposed \dataset dataset.}
    \label{fig:example}
\end{figure*}

\subsection{Multimodal Large Language Models}
Research on MLLMs has progressed from add‑on visual interfaces to tightly unified vision–language architectures. Early adapters such as VisualGPT~\cite{wuVisualChatGPTTalking2023}, which grafts a self-reviving visual encoder onto GPT2 for data-efficient captioning, and GPT-4o~\cite{openaiGPT4TechnicalReport2023}, which simply enables image input for a general-purpose LLM, showed that pre-trained text decoders can address visual tasks. Later work pursues deeper fusion: Flamingo~\cite{alayracFlamingoVisualLanguage2022} bridges frozen vision and language backbones with cross‑attention, while BLIP‑2~\cite{liBLIP2BootstrappingLanguageImage2023} links off‑the‑shelf encoders through a lightweight querying transformer. Leveraging CLIP’s contrastive alignment of image–text embeddings~\cite{radfordLearningTransferableVisual2021}, LLaVA~\cite{liuVisualInstructionTuning2023} feeds CLIP visual tokens directly into a chat‑oriented LLM for unified multimodal reasoning. Scaling this paradigm, Qwen‑VL~\cite{baiQwenVLVersatileVisionLanguage2023} and InternVL~\cite{chenInternVLScalingVision2024} co‑train large vision–language encoders and attain state‑of‑the‑art results across captioning, VQA and grounding. DeepSeek‑VL~\cite{luDeepSeekVLRealWorldVisionLanguage2024} further introduces a hybrid multiscale vision backbone that preserves linguistic fluency while processing high‑resolution images. Collectively, these works chart a clear trend toward instruction‑tuned MLLMs that operate in a shared semantic space across modalities. The \dataset dataset we propose serves as a comprehensive evaluation base for the latest representative MLLMs.


\section{Our \dataset Benchmark}
\subsection{Task Formulation}

The core objective of \dataset{} is to provide a comprehensive framework for evaluating the multimodal physics reasoning capabilities of MLLMs.
As shown in Figure~\ref{fig:example}, each problem instance in \dataset{} is represented by a multimodal input, denoted as $\boldsymbol{M}$. This input $\boldsymbol{M}$ comprises three key components: a Visual Context $\boldsymbol{I}$ (e.g., diagrams, experimental setups), a Textual Context $\boldsymbol{T}$ (e.g., problem descriptions, conditions), and a specific Query $\boldsymbol{Q}$ related to the physics scenario, such that $\boldsymbol{M} = (\boldsymbol{I}, \boldsymbol{T}, \boldsymbol{Q})$.

Given a multimodal input $\boldsymbol{M}_i$ for a problem instance, an MLLM is tasked to generate a structured output that demonstrates its understanding across three key dimensions: Variable Identification, Process Formulation, and Solution Derivation. The model's overall output for an instance $\boldsymbol{M}_i$ can be represented as $\boldsymbol{O}_i = (\boldsymbol{O}_{V,i}, \boldsymbol{O}_{P,i}, \boldsymbol{O}_{S,i})$, corresponding to the outputs for these three dimensions. The ground truth annotations for the same instance are denoted as $\boldsymbol{G}_i = (\boldsymbol{G}_{V,i}, \boldsymbol{G}_{P,i}, \boldsymbol{G}_{S,i})$. The evaluation of the model's output $\boldsymbol{O}_i$ against the ground truth $\boldsymbol{G}_i$ is performed by a judge function $J(\cdot, \cdot)$, implemented using GPT-4o.

For \textbf{Variable Identification}, the model is required to identify $N_V=6$ predefined categories of physical variables from the input $\boldsymbol{M}_i$. The model's output for this subtask is $\boldsymbol{O}_{V,i} = \{o_{v,1}, o_{v,2}, \dots, o_{v,N_V}\}$, where each $o_{v,j}$ corresponds to one of the following components: (1)\textbf{Entity}, (2)\textbf{Geometry}, (3)\textbf{Field}, (4)\textbf{Structure}, (5)\textbf{Connection}, and (6)\textbf{External Influence}. The ground truth is $\boldsymbol{G}_{V,i} = \{g_{v,1}, g_{v,2}, \dots, g_{v,N_V}\}$. Each identified component $o_{v,j}$ is compared with its corresponding ground truth $g_{v,j}$ by the judge, which assigns a boolean score $s_{v,j} = J(o_{v,j}, g_{v,j}) \in \{\textsc{True}, \textsc{False}\}$.

For \textbf{Process Formulation}, the model must describe the physical process by formulating $N_P=5$ types of descriptors. The model's output for this subtask is $\boldsymbol{O}_{P,i} = \{o_{p,1}, o_{p,2}, \dots, o_{p,N_P}\}$, where each $o_{p,k}$ corresponds to one of the following descriptors: (1)\textbf{Entity State}, (2)\textbf{Process Detail}, (3)\textbf{Force \& Energy}, (4)\textbf{State Change}, and (5)\textbf{Process Relation}. The ground truth is $\boldsymbol{G}_{P,i} = \{g_{p,1}, g_{p,2}, \dots, g_{p,N_P}\}$. Each formulated descriptor $o_{p,k}$ is compared against its ground truth $g_{p,k}$ by the judge, which assigns a boolean consistency score $s_{p,k} = J(o_{p,k}, g_{p,k}) \in \{\textsc{True}, \textsc{False}\}$.

For \textbf{Solution Derivation}, the model is required to generate a detailed, step-by-step reasoning chain $\boldsymbol{O}_{S,i}$ that leads to the final answer for the query $\boldsymbol{Q}$ in the input $\boldsymbol{M}_i$. The ground truth is a reference step-by-step solution $\boldsymbol{G}_{S,i}$. The model's generated solution $\boldsymbol{O}_{S,i}$ is compared with the ground truth solution $\boldsymbol{G}_{S,i}$ for logical coherence and correctness of each step by the judge, which assigns an overall boolean score $s_{S,i} = J(\boldsymbol{O}_{S,i}, \boldsymbol{G}_{S,i}) \in \{\textsc{True}, \textsc{False}\}$ based on exact agreement of the entire reasoning chain.

The performance on each dimension is quantified using accuracy metrics. The accuracy for Variable Identification, $\text{Accuracy}_V$, is calculated as the proportion of correctly identified components:
\vspace{-4mm}
\begin{equation}
    \text{Accuracy}_V = \frac{1}{N_V} \sum_{j=1}^{N_V} \mathbb{I}(s_{v,j} = \textsc{True}),
\end{equation}
where $\mathbb{I}(\cdot)$ is the indicator function. Similarly, the accuracy for Process Formulation, $\text{Accuracy}_P$, is
\begin{equation}
    \text{Accuracy}_P = \frac{1}{N_P} \sum_{k=1}^{N_P} \mathbb{I}(s_{p,k} = \textsc{True}).
\end{equation}
The accuracy for Solution Derivation, $\text{Accuracy}_S$, is directly given by
\begin{equation}
    \text{Accuracy}_S = \mathbb{I}(s_{S,i} = \textsc{True}).
\end{equation}

This multi-dimensional task formulation allows \dataset{} to comprehensively assess an MLLM's ability to not only predict a final answer but also to understand the underlying physical variables and processes involved in addressing the query $\boldsymbol{Q}$ based on the multimodal context $(\boldsymbol{I},\boldsymbol{T})$.

\subsection{Data Preparation \& Enhancement}
\begin{figure*}[t!]
    \centering
    \includegraphics[width=0.8\textwidth]{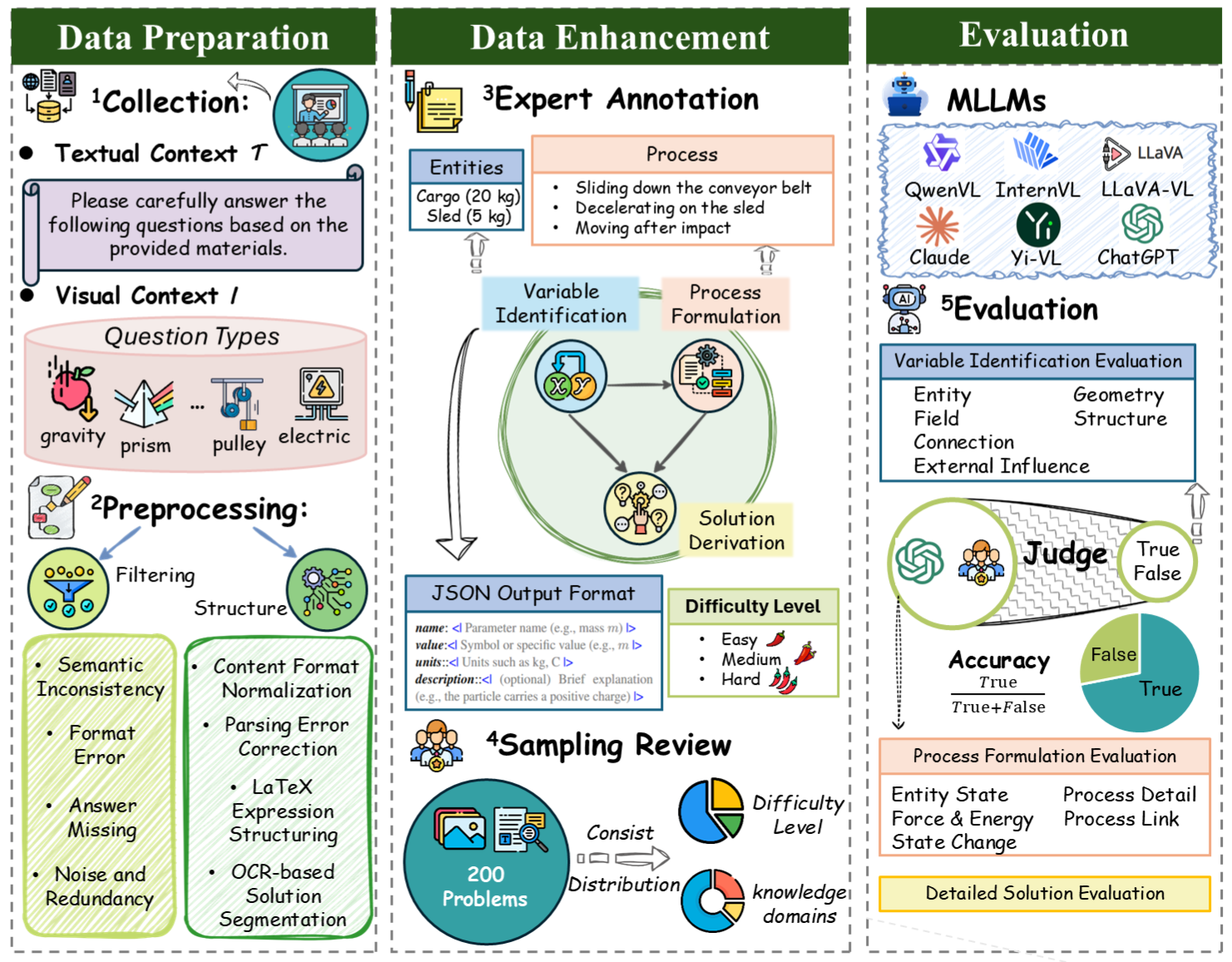}
    \caption{Roadmap of \dataset dataset preparation, enhancement, and evaluation.}
    \label{fig:roadmap}
    \vspace{-4mm}
\end{figure*}

The construction of the \dataset benchmark is a meticulous multi-stage process, designed to ensure the dataset's quality and utility for multimodal physics reasoning. This comprehensive endeavor encompasses four primary stages: initial data collection, rigorous preprocessing, AI-assisted expert annotation, and a final meticulous sampling review (details in Appendix~\ref{app:dataset_preparation}).

\paragraph{Data Collection}
We systematically gathered diverse high-school physics problems, employing custom Python spiders to harvest textual components (stems, options, solutions, answers) and associated visual materials (diagrams, formula images). This collection supports the benchmark's multimodal nature, encompassing various question types like those involving gravity, prisms, \textit{etc}.

\paragraph{Preprocessing}
Raw data underwent extensive preprocessing, including HTML cleaning with regular expressions and GPT-4o, and OCR for formula images to reconstruct LaTeX expressions. This rigorous filtering and structuring addressed inconsistencies and errors, excluded declarative knowledge items, and removed low-quality images, ensuring data integrity and a focus on procedural reasoning.

\paragraph{Expert Annotation}
Cleaned and structured data was enriched through expert annotation, leveraging GPT-4o with designed prompts (see Appendix~\ref{app:task_prompts}) to automatically generate detailed JSON annotations for each problem. These annotations specified relevant variables (entities, properties, values/units) and the formulation of physical processes, and assigned a difficulty level (Easy, Medium, Hard) to each problem.

\paragraph{Sampling Review}
Finally, a stratified subset of 200 items, reflecting original distributions of knowledge domains and difficulty, was selected for thorough manual review. Human experts meticulously examined these items to verify the accuracy of annotations, particularly for variable identification and process formulation, and to ensure the quality and reliability of \dataset.

\subsection{Dataset Details}
\begin{table}[t]
\centering
\footnotesize
\renewcommand\tabcolsep{4pt}
\renewcommand\arraystretch{1.1}
\begin{tabular}{lc}
    \toprule
    \textbf{Statistics} & \textbf{Number} \\
    \midrule
    \textbf{Total Questions} & 5,103 \\
    \midrule
    \textbf{Difficulty Levels} &  \\
    ~- Easy & 2,077 (40.7\%) \\
    ~- Medium & 1,847 (36.2\%) \\
    ~- Hard & 1,179 (23.1\%) \\
    \midrule
    \textbf{Topics} &  \\
    ~- Magnetic Field & 1,080 (21.2\%) \\
    ~- Electromagnetic Induction & 1,063 (20.8\%) \\
    ~- Newton's Laws of Motion & 1,012 (19.8\%) \\
    ~- Electrostatic Field & 642 (12.6\%) \\
    ~- Curvilinear Motion & 526 (10.3\%) \\
    ~- Interaction & 296 (5.8\%) \\
    ~- Conservation of Momentum & 145 (2.8\%) \\
    ~- Uniformly Accelerated Linear Motion & 117 (2.3\%) \\
    ~- Gravitation \& Spaceflight & 67 (1.3\%) \\
    ~- Alternating Current & 56 (1.1\%) \\
    ~- Direct Current & 41 (0.8\%) \\
    ~- Conservation of Mechanical Energy & 35 (0.7\%) \\
    ~- Mechanical Vibrations \& Waves & 13 (0.3\%) \\
    ~- Description of Motion & 10 (0.2\%) \\
    \bottomrule
\end{tabular}
\caption{Key statistics of the \textsc{PhysicsArena} dataset, including diverse difficulty levels and topics.}
\label{tab:physicsarena_statistics}
\vspace{-4mm}
\end{table}
The \dataset dataset, as summarized in Table~\ref{tab:physicsarena_statistics}, encompasses a total of 5,103 multimodal physics problems. These problems are distributed across three distinct difficulty levels: Easy (40.7\%, 2,077 items), Medium (36.2\%, 1,847 items), and Hard (23.1\%, 1,179 items), ensuring a comprehensive range of challenges. The dataset further exhibits broad topical coverage, with significant representation from areas such as Magnetic Fields (21.2\%), Electromagnetic Induction (20.8\%), and Newton’s Laws of Motion (19.8\%), alongside a diverse array of other fundamental physics concepts. The sample problems are presented in Appendix~\ref{app:problem_samples}.

\section{Experiments and Analysis}
\subsection{Evaluation Protocols}
We employ \textbf{GPT-4o} as the automatic judge for \dataset. The evaluation is divided into three complementary subtasks that together assess the \emph{structural} and \emph{procedural} quality of a model's physical reasoning. See details of evaluation protocols and prompts in Appendix~\ref{app:eval_protocol} and~\ref{app:judgement_prompt}.

\subsection{Experimental Setup}
We conduct a comprehensive evaluation on a diverse set of state-of-the-art MLLMs. Our open-source set includes \textbf{InternVL~2.5} series~\cite{Chen2025InternVL25}, \textbf{Qwen~2.5-VL} series~\cite{Bai2025Qwen25VL}, \textbf{LLaVA~v1.6}~\cite{Liu2023LLaVA}, and \textbf{Yi-VL-6B}~\cite{01AI2024Yi}. For closed source, we evaluate three leading models: \textbf{GPT-4o}~\cite{openaiGPT4TechnicalReport2023}, \textbf{Claude~3.5~Sonnet}~\cite{Anthropic2024Claude} and \textbf{Qwen-VL-Max}~\cite{QwenVLMax2024}.

\subsection{Experimental Analysis}
\subsubsection{Main Results}

\begin{figure*}[ht!]
    \centering
    \includegraphics[width=0.8\textwidth]{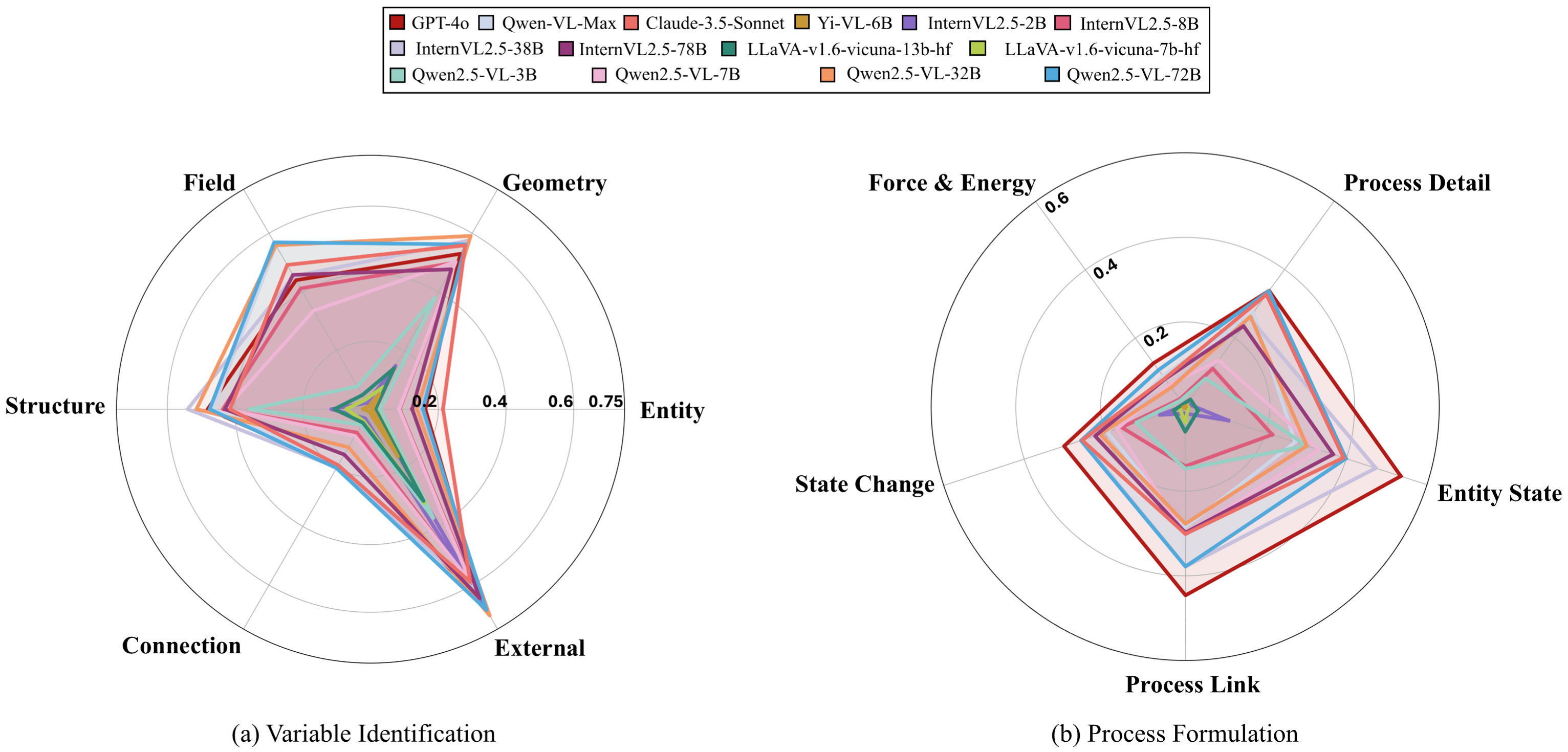}
    \caption{Performance comparison for \textit{Variable Identification} (a) and \textit{Process Formulation} (b).}
    \label{fig:rador_variable_and_process}
    \vspace{-4mm}
\end{figure*}


\begin{table}[htbp]
  \footnotesize
  \renewcommand{\arraystretch}{1.2}
  \setlength{\tabcolsep}{2pt}      
  \centering
  \begin{tabularx}{\columnwidth}{@{\extracolsep{\fill}} l  
      c c c}                            
    \toprule
    \textbf{Model} & \textbf{LLM Base} & \textbf{ViT Encoder} & \textbf{Accuracy (\%)} \\ 
    \midrule
    \multicolumn{4}{c}{\textit{Open-source MLLMs}} \\
    \midrule
    InternVL2.5-2B           & Int2.5-1.8B & IntViT-300M & 3.02  \\
    InternVL2.5-8B           & Int2.5-7B   & IntViT-300M & 9.90  \\
    InternVL2.5-38B          & Q2.5-32B    & IntViT-6B   & 22.95 \\
    Intern2.5VL-78B          & Q2.5-72B    & IntViT-6B   & 21.16 \\
    Qwen2.5-VL-3B            & Q2.5-3B     & Q2ViT-600M  & 8.39  \\
    Qwen2.5-VL-7B            & Q2.5-7B     & Q2ViT-600M  & 14.38 \\
    \textbf{Qwen2.5-VL-32B}  & \textbf{Q2.5-32B} & \textbf{Q2ViT-0.6B} & \textbf{30.59} \\
    Qwen2.5-VL-72B           & Q2.5-72B    & Q2ViT-600M  & 30.49 \\
    Yi-VL-6B                 & Yi-6B       & ViT-H-630M & 0.06  \\
    LLaVA-v1.6-7B            & Vic-7B      & ViT-L-0.43B & 0.37  \\
    LLaVA-v1.6-13B           & Vic-13B     & ViT-L-0.43B & 0.20  \\
    \midrule
    \multicolumn{4}{c}{\textit{Closed-source MLLMs}} \\
    \midrule
    \textbf{Qwen-VL-Max}              & - & - & \textbf{33.47} \\
    GPT-4o                   & - & - & 20.71 \\
    Claude-3.5-Sonnet        & - & - & 23.99 \\
    \bottomrule
  \end{tabularx}
  \caption{Solution derivation accuracy (\%) performance. Abbreviations: \textbf{Int2.5}: InternLM 2.5; \textbf{IntViT}: InternViT; \textbf{Q2.5}: Qwen 2.5; \textbf{Q2ViT}: Qwen2ViT; \textbf{ViT-H/L}: CLIP ViT-H/14 or ViT-L/14; \textbf{Vic}: Vicuna.}
  \label{tab:accuracy_all}
  \vspace{-4mm}
\end{table}

\begin{figure}[th]
  \centering
  \includegraphics[width=\linewidth]{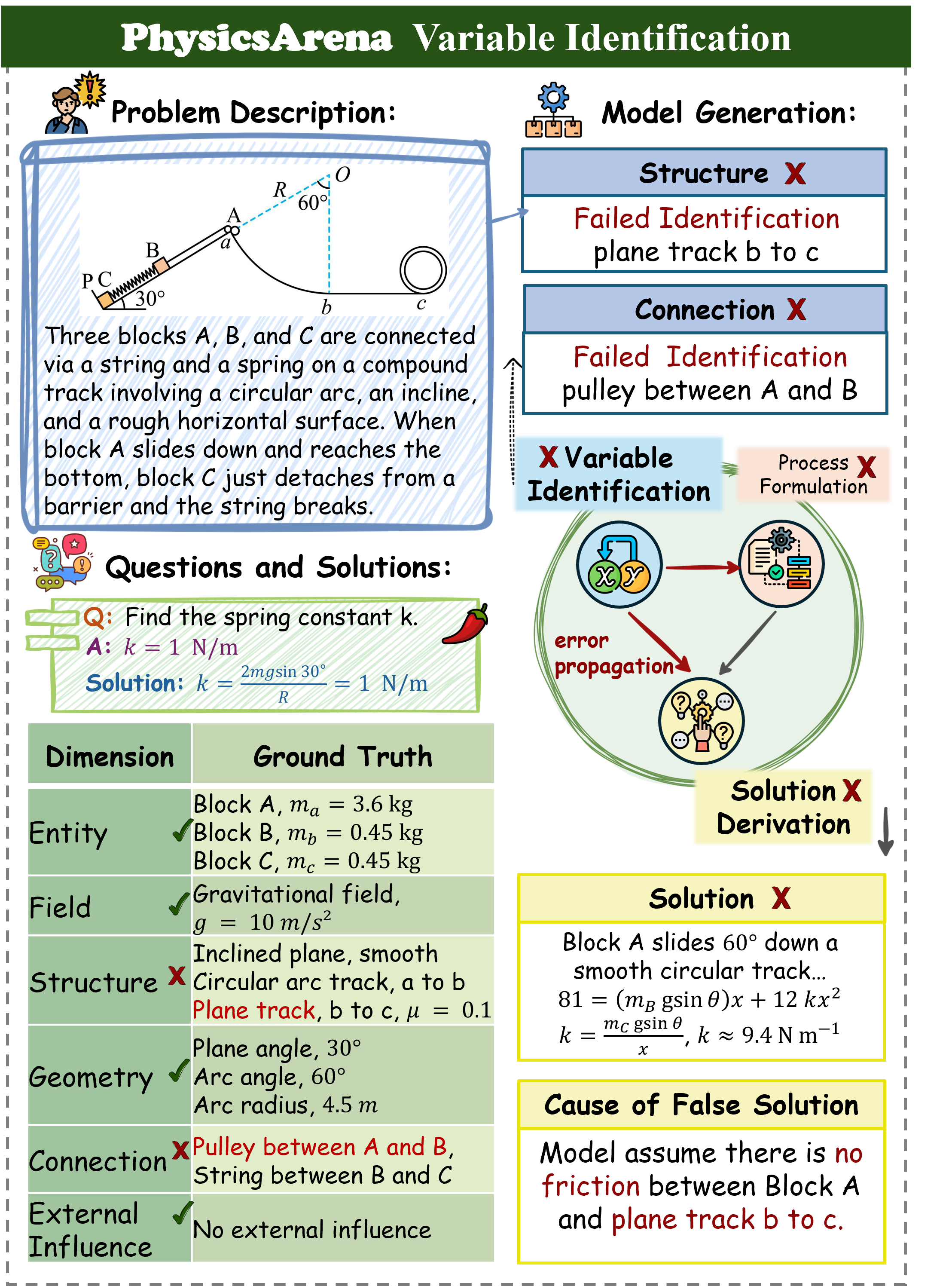}
  \caption{Illustration of a representative bad case of variable identification (more cases in Appendix~\ref{app:case_study}).}
  \label{fig:bad-case-variable-1}
  \vspace{-4mm}
\end{figure}

\textbf{Across all three tasks, the accuracies of state‑of‑the‑art MLLMs remain modest, underscoring the difficulty of the \dataset benchmark.}
In \textit{Variable Identification}, the highest score on any sub‑metric is only 0.704, attained by Qwen2.5‑VL‑32B‑Instruct on \textit{External Influences} (Figure~\ref{fig:rador_variable_and_process}\,(a)); every other dimension lies well below~0.70.  
Models are relatively stronger on \textit{Field}, \textit{Structure}, and \textit{Geometry}, probably because these attributes are stated explicitly in both problem text and accompanying diagram.  
The high numbers for \textit{External Influences} arise because most high‑school problems do \emph{not} involve external agents, turning it into an easy negative class.  
By contrast, categories that hinge on subtle scene understanding and deeper reasoning—\textit{Entity} and, in particular, \textit{Connection}—show the lowest accuracies.

For \textit{Process Formulation} (Figure~\ref{fig:rador_variable_and_process}\,(b)), no model exceeds 0.535 on any metric; the top score (0.535) is achieved by GPT‑4o on \textit{Entity State}.  
While models can enumerate entities, sketch coarse \textit{Process Links}, and provide partial \textit{Process Details}, they struggle with fine‑grained \textit{State Change} descriptions and the associated \textit{Force \& Energy} analyses—both essential for rigorous physical reasoning.

\textit{Solution Derivation} (Table~\ref{tab:accuracy_all}) is the most challenging stage: the best accuracy, 0.335, belongs to Qwen‑VL‑Max. 
The monotonic drop from \textit{Variable Identification} through \textit{Process Formulation} to \textit{Solution Derivation} mirrors the cognitive steps of human problem solving and confirms the progressive difficulty embedded in \dataset.

\textbf{Larger MLLMs consistently outperform smaller ones, and among open‑source systems the Qwen2.5‑VL family leads, followed by InternVL; proprietary Claude and GPT‑4o trail slightly behind.}  
Qwen‑VL‑Max (undisclosed size) attains the highest overall accuracy, while its open‑source siblings Qwen2.5‑VL‑32B‑Instruct and Qwen2.5‑VL‑72B‑Instruct occupy the next two spots. Interestingly, although GPT‑4o lags behind Qwen and Intern on \textit{Variable Identification}, it \emph{tops all five metrics} in \textit{Process Formulation}.  
Thus, stronger low‑level vision grounding benefits \textit{Solution Derivation}, yet the modest ceiling in \textit{Process Formulation} ultimately limits final accuracy.

\textbf{Insufficient visual understanding remains the primary bottleneck for physics reasoning in current MLLMs.}  
Although GPT‑4o leads every sub‑metric in \textit{Process Formulation}, its \textit{Solution Derivation} accuracy is still lower than that of Claude‑3.5‑Sonnet. Despite comparable aggregate scores in \textit{Variable Identification}, Claude surpasses GPT‑4o on the vision‑heavy categories \textit{Entity}, \textit{Geometry}, and \textit{Field}, directly boosting its final‑answer accuracy.  
GPT‑4o nonetheless excels at \textit{Structure} recognition; its weaker performance on physics reasoning stems more from the domain-specific visual grounding demanded by \dataset.

\subsubsection{Bad Case Analysis}




In \textbf{Variable Identification}, models often fail to recognize essential physical components under the problem setting, such as pulleys or springs (see Figure~\ref{fig:bad-case-variable-1}), or hallucinate components that are not present. Another common issue is the misinterpretation of implicit physics scene semantics.

Errors in \textbf{Process Formulation} typically fall into two categories: incorrect process assumptions, such as mistaking circular motion for linear motion, and the omission of key procedural steps, particularly in scenarios involving multiple interacting phases. These errors undermine the model’s ability to construct a coherent and complete internal representation of the physical process, which is critical for successful reasoning.

Notably, failures in \textbf{Solution Derivation} often share the same underlying issues as in the cases.

\subsubsection{Correlation Analysis}
\begin{figure}[t]
  \centering
  \begin{subfigure}[t]{0.40\textwidth}
    \centering
    \includegraphics[width=\linewidth]{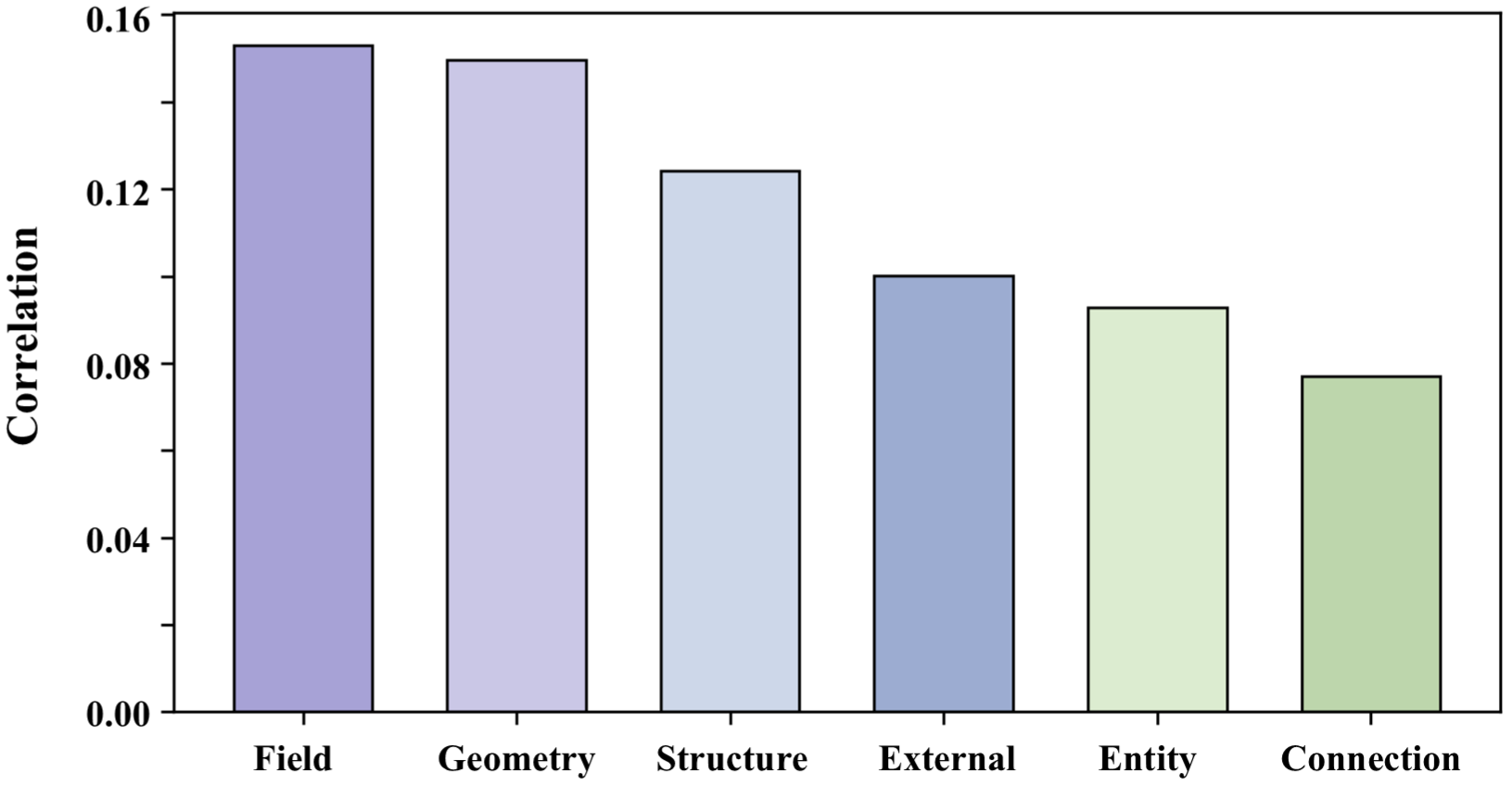}
    \caption{Correlation Between Variable Identification Factors and Solution Accuracy. }
    \label{fig:correlation_variable}
  \end{subfigure}
  \hfill
  \begin{subfigure}[t]{0.40\textwidth}
    \centering
    \includegraphics[width=\linewidth]{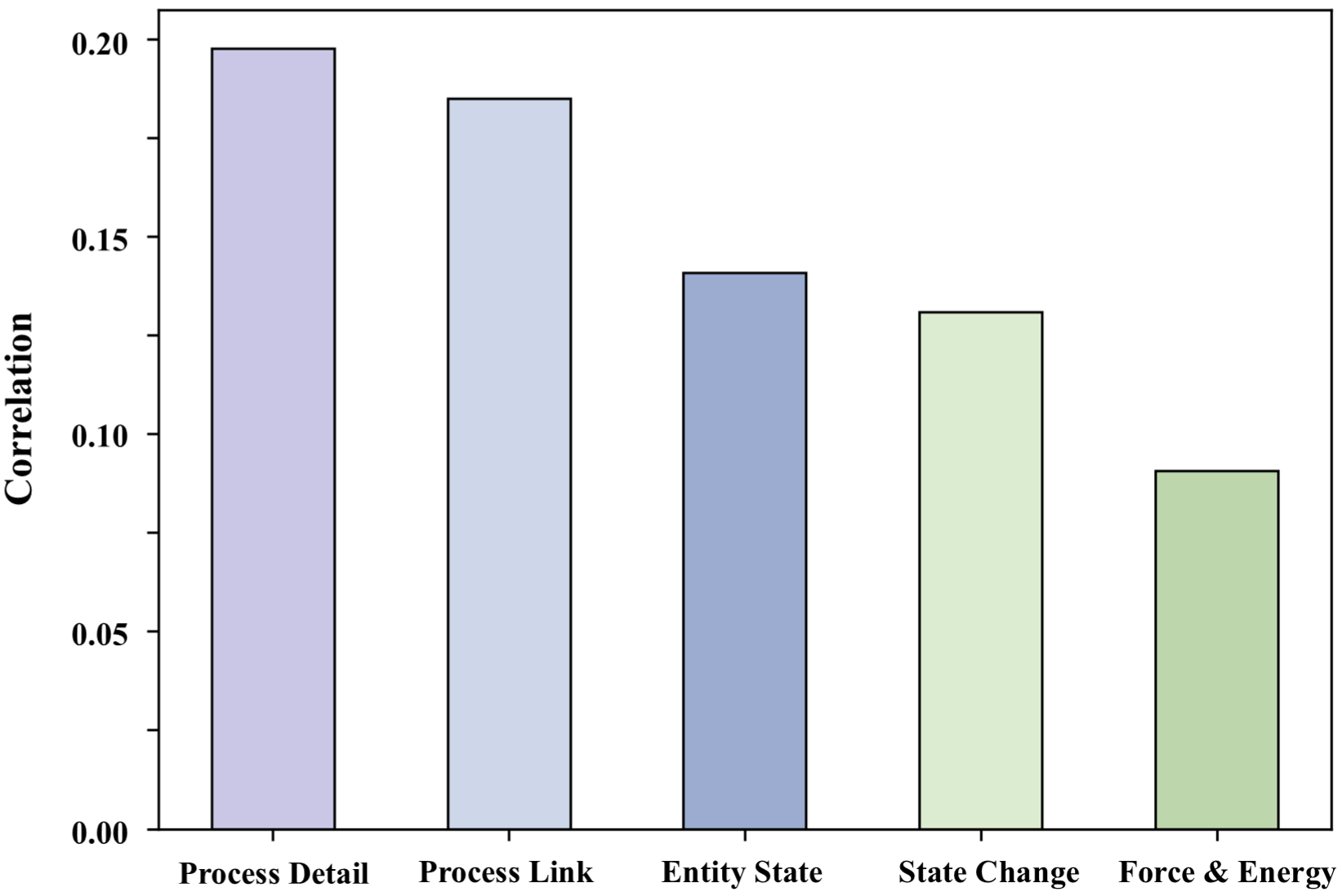}
    \caption{Correlation Between Variable Identification Factors and Solution Accuracy.}
    \label{fig:correlation_process}
  \end{subfigure}
  \caption{Pearson correlation analysis of Variable Identification and Process Formulation factors in relation to Solution Derivation accuracy.}
  \label{fig:correlation}
  \vspace{-4mm}
\end{figure}



We conduct a Pearson correlation analysis~\cite{sedgwick2012pearson} to assess how the correctness of \textbf{Variable Identification} and \textbf{Process Formulation} relates to the accuracy of \textbf{Solution Derivation}. Appendix \ref{app:correlation_analysis} presents a difficulty-level analysis.

The results demonstrate a statistically significant correlation, with $p$-values well below the conventional threshold of $10^{-3}$. As shown in Figure~\ref{fig:correlation_variable}, factors such as \textit{field}, \textit{geometry}, and \textit{structure}—which require effective vision-language alignment to capture physical semantics—exhibit stronger correlations with successful solution derivation in multimodal physics reasoning tasks. 


Similarly, evaluation factors for Process Formulation are also significantly correlated with final solution correctness (with $p$-values well below $10^{-3}$), as shown in Figure~\ref{fig:correlation_process}. This observation is consistent with physical intuition: accurate analysis of procedural details and inter-process dependencies is essential for producing correct solutions in complex multi-step physics problems.



\subsubsection{Difficulty Level Analysis}

\begin{figure}[t]
  \centering
  \includegraphics[width=0.8\linewidth]{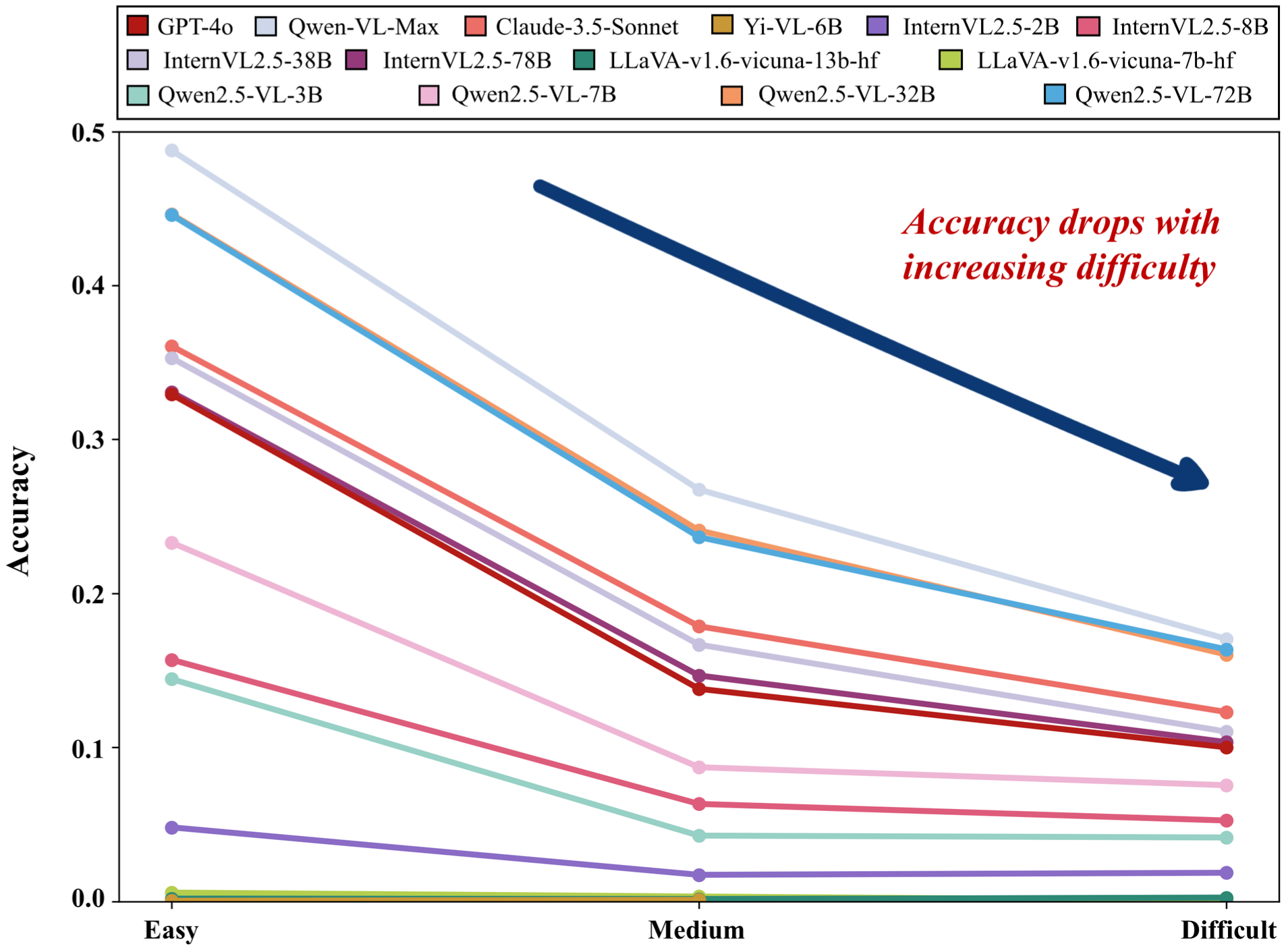}
  \caption{Solution accuracy across difficulty levels.}
  \label{fig:difficulty-analysis}
  \vspace{-4mm}
\end{figure}
 The analysis of model accuracy across different difficulty levels according to Figure~\ref{fig:difficulty-analysis} reveals two clear trends: (1) as task difficulty increases, the overall accuracy of all models declines, and (2) the performance gap between models progressively narrows, indicating a convergence in capabilities.

This convergence suggests a common performance bottleneck faced by current MLLMs when confronted with complex tasks. While models such as the Qwen2.5-VL and InternVL2.5 families demonstrate strong multimodal understanding on easy and medium-level tasks, this advantage diminishes as task complexity grows. At higher difficulty levels, the primary challenge appears to shift from multimodal alignment and semantic understanding to abstract modeling and causal reasoning.

\subsubsection{Scaling Analysis}
\begin{figure}[t]
  \centering
  \includegraphics[width=0.8\linewidth]{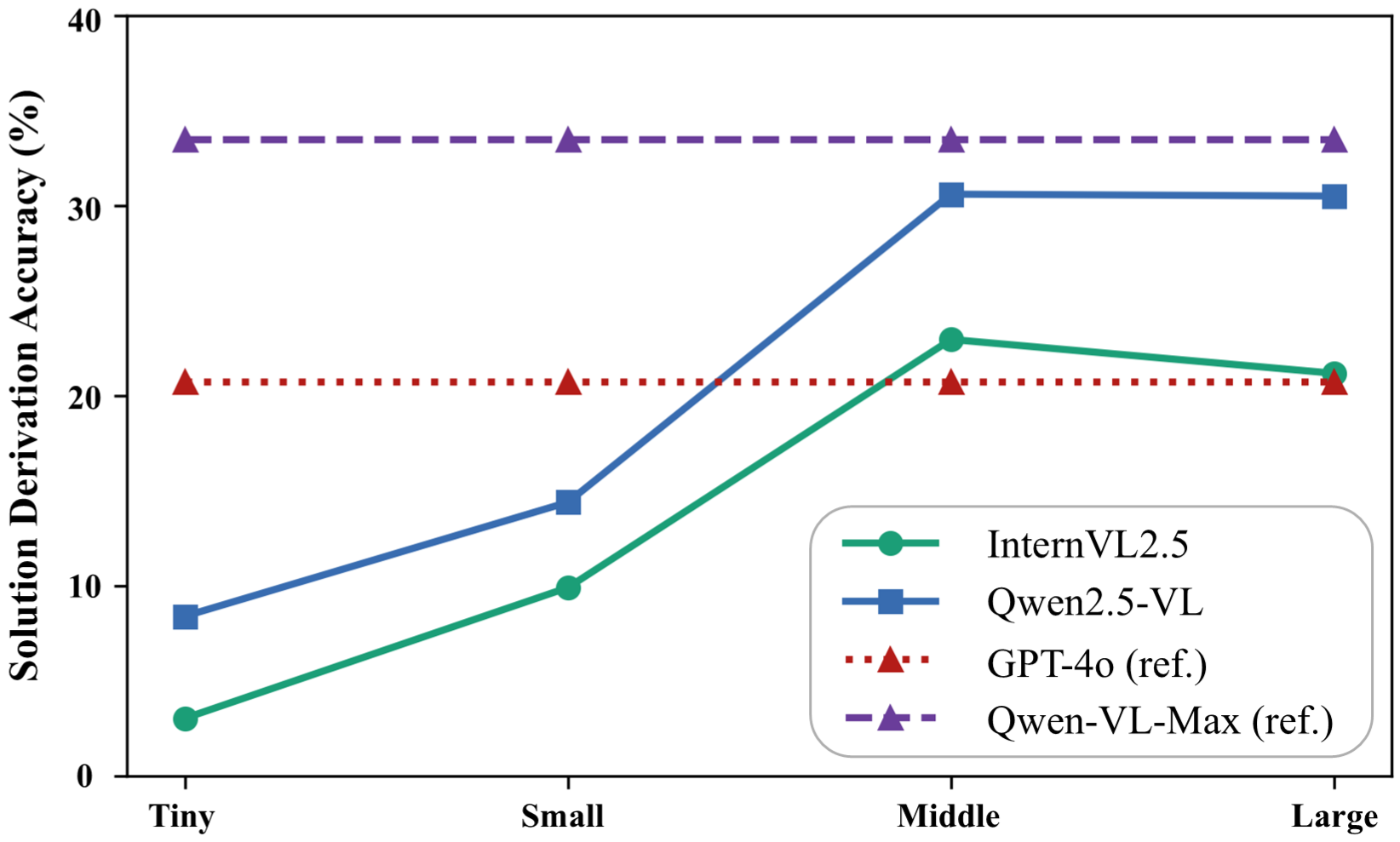}
  \caption{The accuracy of solution derivation of Qwen2.5VL and InternVL2.5. We denote Tiny, Small, Middle, Large as the 2B, 8B, 26B, 78B for InternVL2.5 and 3B, 7B, 32B, 72B for Qwen2.5VL, respectively.}
  \label{fig:solution_scaling_analysis}
  \vspace{-4mm}
\end{figure}
As shown in Figure~\ref{fig:solution_scaling_analysis}, while the accuracy of solution derivation demonstrates a general trend of improvement for both the InternVL2.5 and Qwen2.5-VL models as their size increases from \textit{Tiny} to \textit{Middle}, the accuracy plateaus or even declines when the model size reaches the \textit{Large} scale. We attribute this phenomenon to the challenging nature of \dataset: merely increasing size without task-specific fine-tuning is insufficient. 

According to Table~\ref{tab:accuracy_all}, both InternVL2.5 and Qwen2.5-VL utilize same LLMs in the \textit{Middle} and \textit{Large} settings. However, InternVL2.5 incorporates a 6B InternViT vision encoder, whereas Qwen2.5-VL adopts a unified 600M Qwen2ViT across all scales. Despite the larger parameter size of InternViT, the differing training data and methodologies suggest that Qwen2ViT's training is more efficient~\cite{baiQwen25VLTechnicalReport2025a, chenExpandingPerformanceBoundaries2024}. Furthermore, both models undergo supervised fine-tuning and direct preference optimization in their post-training phases, yet the task settings and training data vary between them. This underscores the importance of fine-tuning in multimodal physical reasoning tasks.
\vspace{-2mm}
\section{Conclusion}
\vspace{-2mm}
We introduced \dataset, the first multimodal physics reasoning benchmark designed to holistically evaluate MLLMs across three critical dimensions: \textit{Variable Identification}, \textit{Process Formulation}, and \textit{Solution Derivation}, with over 5,000 multimodal instances.
Our extensive experiments reveal that while progress has been made, current models still exhibit modest performance, \textit{esp.} process formulation and complex solution derivation, highlighting a significant gap towards AGI-level scientific reasoning~\cite{yan2025position}.

\section*{Limitations}
Despite the comprehensive nature of \dataset and its novel three-dimensional evaluation framework, there are still minor limitations that offer avenues for future work:

\begin{enumerate}
    \item While \dataset covers a broad range of high-school level (CEE equivalent) physics topics, it does not yet extend to more advanced undergraduate or specialized graduate-level physics problems, which often involve more abstract concepts and complex mathematical formalisms. We plan to incrementally expand the dataset to include problems from higher education curricula, thereby increasing the complexity and topical diversity to challenge MLLMs further.

    \item The assessment of Variable Identification and Process Formulation relies on an LLM-based judge (GPT-4o). While scalable and generally effective, automated judges can sometimes miss subtle nuances or exhibit unforeseen biases compared to human expert evaluations, especially for complex reasoning chains. We aim to incorporate periodic, large-scale human expert validation for these intermediate steps and explore hybrid evaluation models that combine the scalability of automated judges with the precision of human oversight.

    \item The current visual inputs in \dataset are primarily static diagrams and images. Real-world physics understanding often involves interpreting dynamic scenarios, such as videos of experiments or interactive simulations. We intend to explore the integration of dynamic multimodal inputs, such as short video clips or simplified interactive environments, to assess MLLMs' ability to reason about temporal changes and cause-and-effect in physical systems.

\end{enumerate}


\bibliography{physicsarena}

\clearpage
\appendix

\section{Details of Data Preparation \& Enhancement}
\label{app:dataset_preparation}
The construction of the \dataset benchmark is a meticulous multi-stage process, designed to ensure the dataset's quality and utility for multimodal physics reasoning. This comprehensive endeavor encompasses four primary stages, as illustrated in Figure~\ref{fig:roadmap}: initial data collection, rigorous preprocessing, sophisticated AI-assisted expert annotation, and a final meticulous sampling review. Each stage builds upon the previous, progressively refining the data towards a high-quality benchmark.

\paragraph{Data Collection}
\textit{First}, the foundational stage involves Data Collection. In this step, we systematically gather a diverse range of high-school physics problems. \textit{Specifically}, custom Python spiders are employed to harvest essential textual components—including problem stems, options, detailed solutions, and correct answers—from various online repositories. \textit{Concurrently}, to support the multimodal nature of our benchmark, associated visual materials, such as problem diagrams, images of formula renderings, and screenshots of solution steps, are also captured, encompassing various question types like those involving gravity, prisms, pulleys, and electric circuits.

\paragraph{Preprocessing}
\textit{Next}, following the initial collection, the raw data undergoes an extensive Preprocessing stage to ensure its integrity and usability. \textit{Initially}, raw HTML content is meticulously cleaned using a combination of regular expressions and a GPT-4o-based corrector; this serves to normalize its structure and accurately extract relevant textual segments. \textit{Subsequently}, any images containing mathematical formulas are processed using OCR to reconstruct their corresponding LaTeX expressions, facilitating machine readability and further analysis. \textit{Furthermore}, a crucial validation step is performed where the final result derived from the provided solution is compared against the labeled correct answer, and any samples exhibiting inconsistencies are discarded. \textit{To maintain a focus on procedural reasoning rather than mere fact recall}, items that solely test declarative knowledge are systematically excluded. \textit{Additionally}, images deemed low-quality or non-compliant with our standards are removed. This rigorous filtering and structuring addresses potential issues such as semantic inconsistency, format errors, missing answers, noise, and redundancy, while also ensuring content format normalization, parsing error correction, LaTeX expression structuring, and effective OCR-based solution segmentation.

\paragraph{Expert Annotation}
\textit{Subsequently}, once the data is cleaned and structured, the Expert Annotation phase commences, aimed at enriching the dataset with crucial reasoning elements. \textit{In this critical phase}, we leverage the advanced capabilities of GPT-4o, guided by carefully designed structured prompts, to automatically generate detailed JSON annotation files for each problem. \textit{These annotations meticulously specify}, as depicted in Figure~\ref{fig:roadmap}, the identification of relevant variables (e.g., entities like "Cargo (20 kg)" or "Sled (5 kg)", their properties, and associated values/units) and the formulation of the physical processes involved (e.g., "Sliding down the conveyor belt," "Decelerating on the sled," "Moving after impact"). The annotation schema is designed to break down the problem into variable identification, process formulation, and ultimately, solution derivation. \textit{Moreover}, each problem is assigned a difficulty level (Easy, Medium, Hard) based on its complexity. 

\paragraph{Sampling Review}
\textit{Finally}, the concluding stage in our data preparation and enhancement pipeline is a thorough Sampling Review to guarantee the accuracy and consistency of the automated annotations. \textit{For this purpose}, we select a stratified subset of 200 items. \textit{This selection is carefully curated} to reflect the original distribution of knowledge domains (\textit{e.g.,} mechanics, electromagnetism, optics) and difficulty tiers within the larger dataset, ensuring the sample's representativeness. \textit{During this stage}, human expert reviewers meticulously examine these selected items. \textit{Their primary focus is twofold}: first, to verify the consistency and correctness of the GPT-4o generated annotations, particularly concerning variable identification and process formulation, and second, to ensure the overall quality and suitability of the problems for the benchmark. This step is crucial for validating the automated annotation process and ensuring the reliability of \dataset.

\section{Task Prompts}
\label{app:task_prompts}
This section outlines the detailed prompt templates used at each stage of the pipeline, including variable identification (Figure~\ref{fig:variable identification prompt}) and process formulation (Figure~\ref{fig:process formulation prompt example}). Each stage is supported by structured JSON formats, as shown in Figures~\ref{fig:json_prompt_full} and~\ref{fig:json_process}, to ensure standardized, machine-readable inputs.

\section{Problem Samples}
\label{app:problem_samples}
\begin{figure*}[ht!]
    \centering
    \begin{tcolorbox}[
        colback=white,
        colframe=black,
        enhanced jigsaw,
        listing only,
        listing options={basicstyle=\rmfamily},
        title={\large Problem Example 1}
    ]
        {\textbf{\textcolor{MidnightBlue}{Problem Description:}}} \\[0.5em]
        Two parallel plates of a capacitor, \(MN\) and \(PQ\), are placed facing each other at an angle of 
        \(\theta = 45^{\circ}\) to the horizontal.  The diagonal \(MQ\) is horizontal.  
        A horizontal insulating guide \(OQ\) is fixed at midpoint \(O\); at its left end two spheres \(A\) and \(B\) are 
        pinned together against a compressed spring.  
        Sphere \(A\) (conducting, charge \(+q\), mass \(m\)) is fired leftward with speed \(v_{0}\) when the pin is released, 
        while sphere \(B\) (insulating, mass \(2m\)) moves rightward on \(OQ\) with kinetic‐friction coefficient \(\mu = \tfrac{1}{16}\). The distances \(OM\) and the plate length equal \(L=\frac{3v_{0}^{2}}{2g}\). Leaving the capacitor at \(M\), sphere \(A\) enters mutually perpendicular uniform electric and magnetic fields 
        and performs uniform circular motion of radius \(\tfrac{4}{\pi}L\); after half a turn it exits horizontally at \(E\) 
        onto a long, smooth, horizontal surface \(EF\).  
        Ignore edge effects; take gravitational acceleration \(g\). \\

        \begin{center}
            \includegraphics[width=0.4\linewidth]{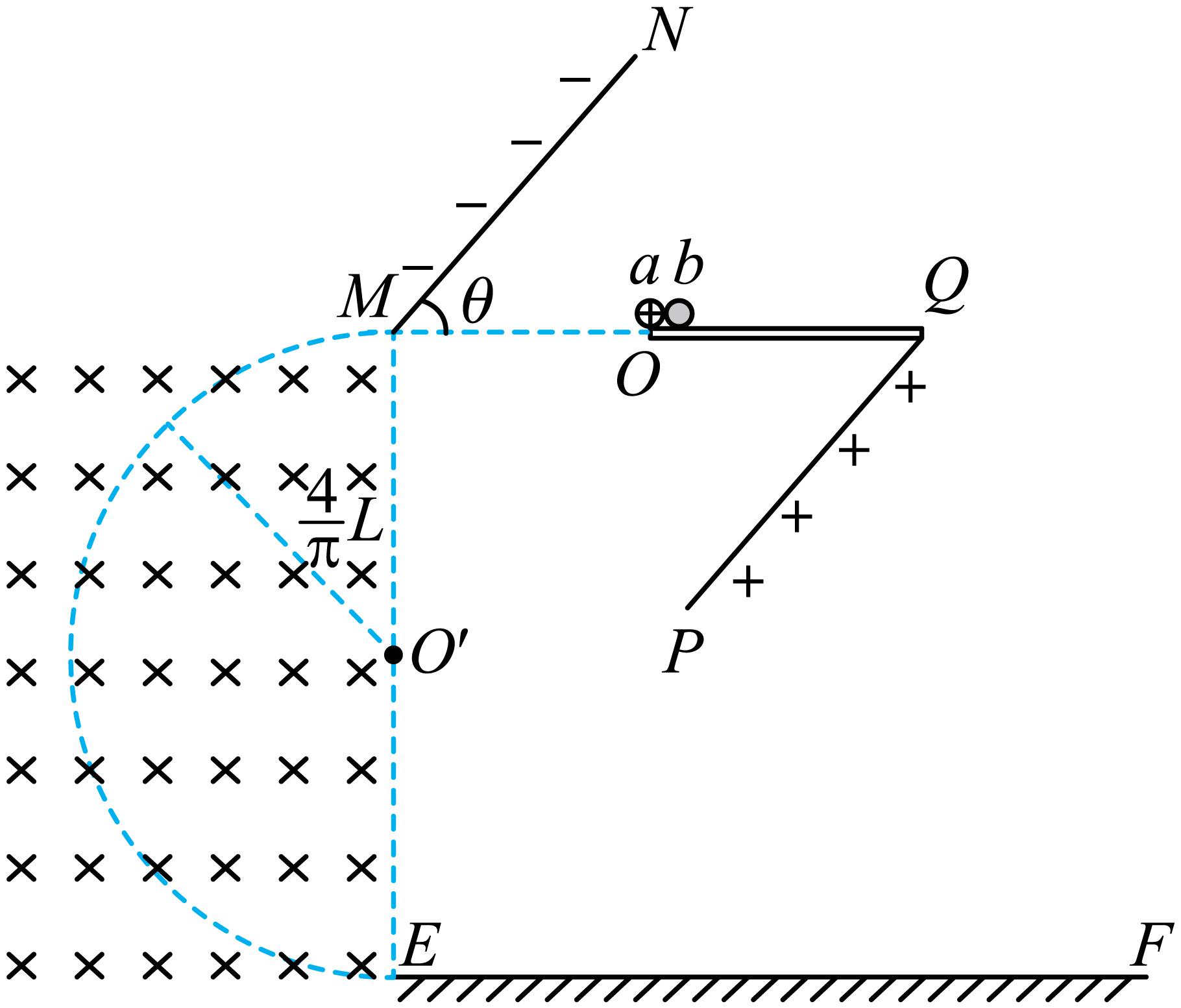}
        \end{center}

        {\textbf{\textcolor{MidnightBlue}{Question:}}} \\
        \textbullet\ Find the speed of sphere \(A\) when it arrives at point \(M\). \\[0.5em]

        {\textbf{\textcolor{MidnightBlue}{Answer:}}} \\
        \[
            v_{M}=2v_{0}.
        \]

        {\textbf{\textcolor{MidnightBlue}{Solution Derivation:}}} \\
            \textbullet\ In the space between plates \(MN\) and \(PQ\) the electric field is uniform, so sphere \(A\) experiences a constant horizontal force \(qE\). Hence its horizontal acceleration is constant, denote it \(a\).\\
            \textbullet\ The horizontal work done by the electric field while \(A\) moves the distance \(OM=L\) equals its gain in kinetic energy:
              \[
                  \tfrac{1}{2}m v_{M}^{2}-\tfrac{1}{2}m v_{0}^{2}=m a L
                  \quad\Longrightarrow\quad
                  v_{M}^{2}=v_{0}^{2}+2aL.
              \]\\
            \textbullet\ Although \(E\) is not stated directly, the given geometric data  
              \(L=\dfrac{3v_{0}^{2}}{2g}\) imply that the acceleration must satisfy 
              \(a=g\) so that the resulting velocity matches subsequent motion constraints.  
              (Indeed, substituting \(a=g\) will yield an integer multiple of \(v_0\).)\\
            \textbullet\ Insert \(a=g\) and the expression for \(L\):
                  \[
                      v_{M}^{2}
                      =v_{0}^{2}+2g\!\left(\frac{3v_{0}^{2}}{2g}\right)
                      =v_{0}^{2}+3v_{0}^{2}=4v_{0}^{2}
                      \;\;\Longrightarrow\;\;
                      v_{M}=2v_{0}.
                  \]
    \end{tcolorbox}
    \caption{Problem Example 1: Problem Description, Question, Answer and Solution Derivation. Variable Identification analysis see Figure~\ref{fig:bad-case-variable-1}.}
    \label{fig:problem_example_1}
\end{figure*}

\begin{figure*}[ht!]
    \centering
    \begin{tcolorbox}[
        colback=white,
        colframe=black,
        enhanced jigsaw,
        listing only,
        listing options={basicstyle=\rmfamily},
        title={\large Problem Example 2}
    ]
        {\textbf{\textcolor{MidnightBlue}{Problem Description:}}}\\[0.5em]
        A small slider \(B\) (mass \(m\)) is placed on a board \(A\) (mass \(M\), length \(L_{0}\)).
        The board rests on a rough incline that forms an angle \(37^{\circ}\) with the horizontal; its
        lower end is a horizontal distance \(x_{0}\) from a rigid stopper \(Q\).  
        Static friction is large enough that \(B\) never slips on \(A\); instead they repeatedly collide
        inelastically with the fixed vertical walls at the two ends of the incline.
        The coefficient of kinetic friction between \(B\) and \(A\) is \(\mu\).
        The system is released from rest at the configuration shown.\\
        
        \begin{center}
            \includegraphics[width=0.6\linewidth]{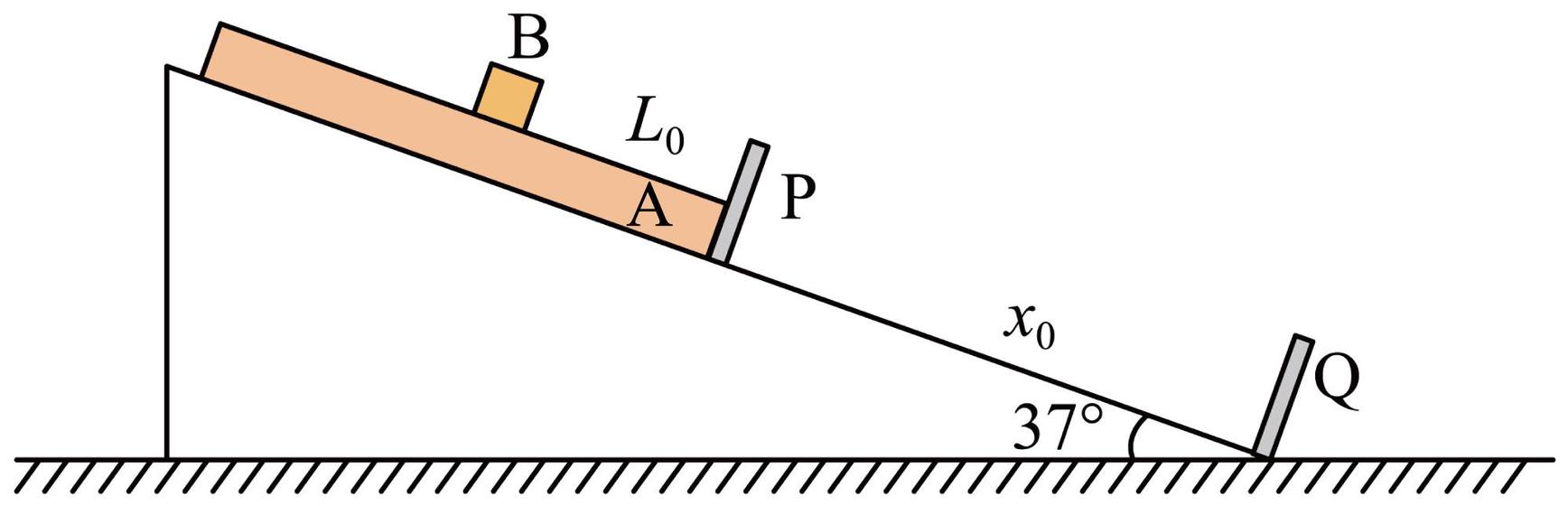}
        \end{center}

        {\textbf{\textcolor{MidnightBlue}{Question Solved:}}}\\[0.5em]
        \textbf{Question:} What is the speed of the slider \(B\) at its \emph{first} collision with the upper wall?\\
        \textbf{Answer:} \(v_{\,\text{B1}} = 3\,\mathrm{m/s}\).\\

        {\textbf{\textcolor{MidnightBlue}{Current Question:}}}\\[0.5em]
         How long does it take for the board \(A\) to collide with the stopper \(Q\)?\\[0.5em]
        {\textbf{\textcolor{MidnightBlue}{Answer:}}}\\[0.5em]
         \(t_{\text{total}} = 2.73\,\mathrm{s}\).\\[0.5em]
        {\textbf{\textcolor{MidnightBlue}{Solution Derivation:}}}\\[0.5em]          
        \textbullet \ Net downslope acceleration of \(B\) while sliding on \(A\):
        \(a = g\sin37^{\circ}-\mu g\cos37^{\circ}\;\,(\approx 6\,\mathrm{m/s^{2}})\).\\[0.2em]
        \textbullet \ \(v_{\,\text{B1}}=at_{1}=3\Rightarrow t_{1}=0.50\,\mathrm{s}\).  
        Speeds exchange: \(v_{\,\text{B1}}'=0,\;v_{\,\text{A1}}'=3\,\mathrm{m/s}\).\\[0.2em]
        \textbullet\ Require \( \tfrac12 at_{2}^{2}=v_{\,\text{A1}}'t_{2}\Rightarrow t_{2}=1.0\,\mathrm{s}\).  Just before contact: \(v_{\,\text{B2}}=at_{2}=6\,\mathrm{m/s}\).  Exchange: \(v_{\,\text{B2}}'=3,\;v_{\,\text{A2}}'=6\,\mathrm{m/s}\).\\
        \textbullet\ Solve \(6t_{3}=3t_{3}+\tfrac12 at_{3}^{2}\Rightarrow t_{3}=1.0\,\mathrm{s}\).  
        \(v_{\,\text{B3}}=9\,\mathrm{m/s}\); exchange gives \(v_{\,\text{A3}}=9\,\mathrm{m/s}\).\\[0.2em]
        \textbullet\ Board has travelled \(x_{1}+x_{2}=3+6=9\,\mathrm{m}\).  
         Remaining distance to stopper: \(\Delta x=x_{0}-9\).  
        Time to cover this at \(v_{\,\text{A3}}=9\,\mathrm{m/s}\):
        \(t_{4}=\dfrac{\Delta x}{9}\).  
        With \(x_{0}=11.07\,\mathrm{m}\), \(t_{4}=0.23\,\mathrm{s}\).\\[0.2em]
        \textbullet\  \(t_{\text{total}}=t_{1}+t_{2}+t_{3}+t_{4}=0.50+1.00+1.00+0.23=2.73\,\mathrm{s}\).
    \end{tcolorbox}
    \caption{Problem Example 2: Problem Description, Question, Answer and Solution Derivation. Process Formulation Analysis see Figure~\ref{fig:bad-case-process-1}.}
    \label{fig: problem_example_2}
\end{figure*}

This section provides two problem samples. See Figure~\ref{fig:problem_example_1} and Figure~\ref{fig: problem_example_2}.

\section{Evaluation Protocol Details}
\label{app:eval_protocol}

\paragraph{Variable Identification Evaluation}\label{sec:var-id}
For each problem instance we extract six components: (1)~\textbf{Entity}—the primary physical entities mentioned; (2)~\textbf{Geometry}—geometric information such as dimensions, shapes, and relative positions; (3)~\textbf{Field}—descriptions of physical fields (gravitational, magnetic, electric); (4)~\textbf{Structure}—fixed, immovable elements (e.g., ground, walls); (5)~\textbf{Connection}—links between entities or between an entity and a structure (e.g., hinges, ropes); and (6)~\textbf{External Influence}—external inputs or hypothesised influences introduced by the problem setter. Each component is compared with the ground truth and labelled \textsc{True} or \textsc{False}.

\paragraph{Process Formulation Evaluation}\label{sec:proc-form}
We model the temporal evolution of the system using five descriptors: (1)~\textbf{Entity State}—the sequence of equilibrium states and dynamic processes for each entity; (2)~\textbf{Process Detail}—preconditions, timestamps, and parameter changes characterising each process; (3)~\textbf{Force \& Energy}—forces acting during each dynamic process and the associated energy transformations; (4)~\textbf{State Change}—the initial and terminal states that bound the dynamic situation; and (5)~\textbf{Process Link}—logical relations between states or processes such as \textit{triggered\_by}, \textit{sequential}, or \textit{simultaneous}. Every descriptor is compared with the ground truth and assigned a Boolean consistency label.

\paragraph{Solution Derivation Evaluation}\label{sec:step-sol}
In addition to the structured representations, we evaluate the model’s \emph{step-by-step} solution. The generated reasoning chain is aligned with the annotated ground truth; exact agreement yields \textsc{True}, while any discrepancy results in \textsc{False}.

\section{Judgement Prompts}
\label{app:judgement_prompt}
To enable automatic evaluation using GPT-4o, we design dedicated judgement prompts for each task stage. These prompts instruct the model to assess the quality and correctness of outputs across variable identification (Figure~\ref{fig:variables_judgement_prompt}), process formulation (Figure~\ref{fig:process_judgement_prompt}), and solution derivation (Figure~\ref{fig:answer_judgement_prompt}), ensuring consistent and reliable evaluation.

\begin{figure*}[ht!]
    \centering
    \begin{tcolorbox}[
        colback=white,
        colframe=black,
        enhanced jigsaw,
        listing only,
        listing options={basicstyle=\rmfamily},
        title={\large Prompt for Variable Identification}
    ]
        {\textbf{\textcolor{MidnightBlue}{Task Definition:}}} You are performing an information extraction task. The current goal is to: extract only all the physical variables and their information that appear in the problem text and diagram (such as names, initial values, units, directions, whether they are known, etc.). Answers and explanation is only for verifying information — you should not introduce extra information from answers and explanations. \\[0.5em]

        {\textbf{\textcolor{MidnightBlue}{Below is the reference content:}}} \\
        Problem Image: <image> \\
        Problem Text: \{text\} \\[0.5em]

        {\textbf{\textcolor{MidnightBlue}{Instruction:}}} Please classify and fill in all the involved physical quantities (for example: mass $m$, charge $q$, velocity $v$, electric field strength $E$, magnetic field strength $B$, etc.) according to the \textbf{JSON template} provided below. If the problem does not clearly specify a particular variable's value, unit, or direction, you may fill in "unknown" or leave it blank. Return only the JSON included by \verb|```|JSON and return only in English. \\[0.5em]

        {\textbf{\textcolor{MidnightBlue}{Note:}}} \\
        1. Do not output the problem's answer, solution process, or derivation explanation. \\
        2. You only need to return text that conforms to the JSON template; do not add any extra text. \\
        3. Keep the field structure and field names consistent with the JSON template; if there is no relevant information, you may leave it blank or remove empty fields.
    \end{tcolorbox}
    \caption{Prompt for Variable Identification.}
    \label{fig:variable identification prompt}
\end{figure*}

\begin{figure*}[ht!]
    \centering
    \begin{tcolorbox}[
        colback=white,
        colframe=black,
        enhanced jigsaw,
        listing only,
        listing options={basicstyle=\rmfamily},
        title={\large Prompt for Process Formulation}
    ]
        {\textbf{\textcolor{MidnightBlue}{Task Definition:}}} You are performing an information extraction task. The goal is to identify and extract all the physical processes (such as motion processes, collision processes, etc.) for each entity (for example, particles, blocks, etc.) described in the problem and diagram. Answers and explanation is only for verifying information, you should not introduce extra information from answers and explanations. \\[0.5em]

        {\textbf{\textcolor{MidnightBlue}{Below is the reference content:}}} \\
        Problem Image: <image> \\
        Problem Text: \{text\} \\[0.5em]

        {\textbf{\textcolor{MidnightBlue}{Instruction:}}} Begin extracting data according to \textbf{JSON template} below, and once finished, return only the JSON included by \verb|```|JSON and return only in English. \\[0.5em]

        {\textbf{\textcolor{MidnightBlue}{Note:}}} \\
        1. Do not provide the problem’s answer, solution steps, or derivations. \\
        2. Only return content related to this information extraction task that aligns with the following JSON template structure. \\
        3. If certain information in the problem is unclear, use "unknown" or omit the corresponding field. \\
        4. Keep the field hierarchy and field names exactly the same as in the template below.
    \end{tcolorbox}
    \caption{Prompt for Process Formulation.}
    \label{fig:process formulation prompt example}
\end{figure*}

\begin{figure*}[ht!]
    \centering
    \begin{tcolorbox}[
        colback=white,
        colframe=black,
        enhanced jigsaw,
        listing only,
        listing options={basicstyle=\rmfamily},
        title={\large Prompt for Solution Derivation}
    ]
        {\textbf{\textcolor{MidnightBlue}{Below is the reference content:}}} \\
        \textbullet\ Problem Image: <image> \\
        \textbullet\ Problem Text: \{text\} \\[0.5em]

        {\textbf{\textcolor{MidnightBlue}{Instruction:}}} Solve the physics problem step by step. Return only in English.
    \end{tcolorbox}
    \caption{Prompt for Solution Derivation.}
    \label{fig:solution derivation prompt example}
\end{figure*}

\begin{figure*}[p]               
  \centering
  \begin{tcolorbox}[colback=white, colframe=black, enhanced jigsaw,
                    title={\large JSON template for Variable Identification (a)}]
  \begin{lstlisting}[language=json]
{
  "entities": [
    {
      "name": "<Entity name>",          // e.g. "charged particle"
      "type": "<Entity type>",          // e.g. "particle"
      "position": "<Position>",         // e.g. "from point a to b"
      "variables": [
        {
          "name": "<Variable name>",    // e.g. "velocity v"
          "value": "<Initial value>",    // e.g. "v_0"
          "units": "<Units>",            // e.g. "m/s"
          "direction": "<Direction>",    // e.g. "upwards"
          "conditions": "<Conditions>",  // e.g. "under gravity"
          "domain": "<Domain>",          // e.g. "$t \in [0, t_b]$"
          "given_or_unknown": "<Known?>" // "known" / "unknown"
        }
        /* ...additional variables... */
      ],
      "parameters": [
        {
          "name": "<Parameter name>",    // e.g. "mass m"
          "value": "<Symbol/value>",     // e.g. "m"
          "units": "<Units>",            // e.g. "kg"
          "description": "<Optional>"
        }
        /* ...additional parameters... */
      ],
      "interactions": "<Interactions>"   // e.g. "subject to E, g"
    }
    /* ...additional entities... */
  ],

  "fields": [
    {
      "name": "<Field name>",           // e.g. "uniform E-field"
      "region": "<Region>",             // e.g. "$x \in [0, L]"$
      "variables": [ ... ],             // e.g. "$E$,$E_0$,N/C,Upward,unknown" 
      "parameters": [ ... ]             // e.g. "vacuum permittivity $\varepsilon_0$" 
    }
    /* ...additional fields... */
  ],

  "structures": [
    {
      "name": "<Structure name>",       // e.g. "metal rail"
      "position": "<Location / Size>",  // e.g. "along x, length L"
      "constants": [
        {
          "name": "<Constant>",         // e.g. "length L"
          "value": "<Symbol/value>",
          "description": "<Optional>"
        }
      ]
    }
    /* ...additional structures... */
  ]

  /*   see Fig (b) for remaining blocks  */
}
  \end{lstlisting}
  \end{tcolorbox}

  \caption{JSON prompt template for variable identification (a): entity, field and structure blocks.}
  \label{fig:json_prompt_full}
\end{figure*}

\begin{figure*}[p]
  \ContinuedFloat          
  \centering
  \begin{tcolorbox}[colback=white, colframe=black, enhanced jigsaw,
                    title={\large JSON template for Variable Identification (b)}]
  \begin{lstlisting}[language=json]
{
  "geometries": [
    {
      "description": "<Geometric relation>", // e.g. "displacement AB"
      "value": "<Symbol / Value>"            // e.g. "$d$" or "$\theta$"
    }
    /* ...additional geometries... */
  ],

  "connections": [
    {
      "description": "<Entity $\,\leftrightarrow\,$ Field interaction>",
      "variables": [ ... ],                  // e.g. "net force F, qE - mg, N"
      "constants": [ ... ]                   // e.g. "friction coefficient, $\mu$"
    }
    /* ...additional connections... */
  ],

  "external_influences": [
    {
      "description": "<External force / circuit>", // e.g. "additional resistor"
      "variables": [ ... ],
      "constants": [ ... ]
    }
    /* ...additional external influences... */
  ]
}
  \end{lstlisting}
  \end{tcolorbox}

  \caption{JSON prompt template for variable identification (b): geometry, interaction and external-influence blocks (continuation of Fig.~\ref{fig:json_prompt_full}).}
\end{figure*}

\begin{figure*}[p]
  \centering
  \begin{tcolorbox}[colback=white, colframe=black, enhanced jigsaw,
                    title={\large JSON template for Process Formulation (a)}]
  \begin{lstlisting}[language=json]
{
  "entities": [
    {
      "id": "<Entity ID>",                 // e.g. "A"
      "name": "<Readable name>",           // e.g. "Block A"

      "situations": [

        /* ------- equilibrium example ------- */
        {
          "situation_id": "<ID_S1>",        // e.g. "A_S1"
          "state_type": "equilibrium",
          "force_balance": "<Equation>",    // e.g. "N = mg"
          "energy_balance": "<Statement>",  // e.g. "No net energy change"
          "additional_info": "<Optional>"
        },

        /* -------- dynamic example --------- */
        {
          "situation_id": "<ID_S2>",        // e.g. "A_S2"
          "state_type": "dynamic",
          "process_name": "<Process>",      // e.g. "Collision with bullet"
          "trigger": "<Trigger>",           // e.g. "Bullet contacts A"
          "start_condition": "<Start>",     // e.g. "A at rest"
          "end_condition": "<End>",         // e.g. "A & bullet move together"
          "process_description": "<Brief description>",
          "forces_involved": [
            {
              "type": "<Force type>",       // e.g. "contact force"
              "magnitude": "<Expression>",  // e.g. "$k \cdot x$"
              "direction": "<Direction>"    // e.g. "horizontal"
            }
          ],
          "energy_transfers": [
            {
              "type": "<Energy type>",
              "description": "<Explanation>"
            }
          ],
          "initial_physical_state": {
            "position": "<Pos>",            // e.g. "$x = 0$"
            "velocity": "<Vel>",            // e.g. "$v = 0$"
            "acceleration": "<Acc>",        // e.g. "$a = 0$"
            "energy": "<Energy>"            // e.g. "KE = 0"
          },

          "final_physical_state": { ... },  // e.g. "x=x_1, v  \neq 0"
          "time_description": "<Duration>"  // e.g. "very short collision"
        }
      ]
    }
    /* ...additional situations... */
  ],
  /* ...additional entities... */
  /*    see Fig. (b) for process relations   */
}
  \end{lstlisting}
  \end{tcolorbox}

  \caption{JSON prompt template for process formulation (a): entity block with two sample situations (equilibrium and dynamic). }
  \label{fig:json_process}
\end{figure*}

\begin{figure*}[p]
  \ContinuedFloat
  \centering
  \begin{tcolorbox}[colback=white, colframe=black, enhanced jigsaw,
                    title={\large JSON template for Process Formulation (b)}]
  \begin{lstlisting}[language=json]
{
  "process_relations": [
    {
      "process_id": "<Proc ID>",                  // e.g. "A_S2"
      "related_processes": ["<Other ID>"],        // e.g. ["B_S1"]
      "relation_type": "<Relation>"               // e.g. "sequential"
    }
    /* ...additional relations... */
  ]
}
  \end{lstlisting}
  \end{tcolorbox}
  \caption{JSON prompt template for process formulation (b): relationship block (continuation of Fig.~\ref{fig:json_process}).}
\end{figure*}

\begin{figure*}[ht!]
    \centering
    \begin{tcolorbox}[
        colback=white,
        colframe=black,
        enhanced jigsaw,
        listing only,
        listing options={basicstyle=\rmfamily},
        title={\large Judgement Prompt for Variable Identification}
    ]   
        {\textbf{\textcolor{MidnightBlue}{Instructions:}}} \\[0.5em]
        You are given two sets of extracted information describing the same physics scenario: \\
        1. \textbf{Ground Truth} — the reference answer (in JSON format) \\
        2. \textbf{Large Language Model} — the model's prediction (in JSON format) \\[0.5em]
        Your task is to evaluate whether they align across the six aspects below. Assign a judgement of \textbf{True} or \textbf{False} based on the following guidelines: \\[0.5em]
        \textbullet\ Mark \textbf{True} if minor wordings or variations in phrasing (e.g., “charged particle” vs. “particle”) \\
        \textbullet\ Mark \textbf{True} if additional but non-conflicting information \\
        \textbullet\ Mark \textbf{False} if missing or extra elements, mismatched values, inconsistent units, or known/unknown status \\
        \textbullet\ Mark \textbf{False} if contradictory or scenario-irrelevant content \\[0.5em]

        {\textbf{\textcolor{MidnightBlue}{Evaluation Aspects:}}} \\[0.5em]
        The comparison should be conducted across the following six aspects. Any variables associated with each aspect (e.g., names, values, units, directions, known/unknown status) should be evaluated as part of that category: \\[0.5em]
        
        \begin{tabularx}{\textwidth}{@{}l X@{}}
        \textbullet\  \textbf{Entity} & Physical objects and their properties (e.g., mass, charge, velocity) \\
        \textbullet\  \textbf{Field} & Any physical fields present (e.g., electric, magnetic) and associated quantities \\
        \textbullet\ \textbf{Structure} & Fixed elements or boundaries (e.g., rails, frames, spatial constraints) \\
        \textbullet\  \textbf{Geometry} & Geometric features and relationships (e.g., lengths, angles, positions) \\
        \textbullet\  \textbf{Connection} & Physical interactions (e.g., forces, constraints, contact conditions) \\
        \textbullet\  \textbf{External Influences} & Externally imposed factors (e.g., applied fields, switching conditions) \\
        \end{tabularx} \\[0.5em]

        {\textbf{\textcolor{MidnightBlue}{Input Format:}}} \\
        \textbullet\ \textbf{Ground Truth}: \{ground\_truth\} \\
        \textbullet\ \textbf{Large Language Model}: \{large\_language\_model\_result\} \\[0.5em]

        {\textbf{\textcolor{MidnightBlue}{Output Format (JSON Template):}}}
\begin{verbatim}
{
  "entity": <boolean>,                 // e.g., True
  "field": <boolean>,                  // e.g., False
  "structure": <boolean>,              // e.g., True
  "geometry": <boolean>,               // e.g., True
  "connection": <boolean>,             // e.g., False
  "external_influences": <boolean>     // e.g., True
}
\end{verbatim}
    \end{tcolorbox}
    \caption{Evaluation prompt used for judging alignment between MLLM-predicted Variable Identification result and ground truth across key physical factors.}
    \label{fig:variables_judgement_prompt}
\end{figure*}

\begin{figure*}[ht!]
    \centering
    \begin{tcolorbox}[
        colback=white,
        colframe=black,
        enhanced jigsaw,
        listing only,
        listing options={basicstyle=\rmfamily},
        title={\large Judgement Prompt for Process Formulation}
    ]        
        {\textbf{\textcolor{MidnightBlue}{Instructions:}}} \\
        You are given two sets of extracted information from the same physics scenario: \\
        1. \textbf{Ground Truth} — the reference answer (in JSON format) \\
        2. \textbf{Large Language Model} — the model's prediction (in JSON format) \\[0.5em]
        Your task is to evaluate whether they align across the five aspects below. Assign a judgement of \textbf{True} or \textbf{False} based on the following guidelines: \\[0.5em]
        \textbullet\ Mark \textbf{True} if minor wording differences (e.g., “impact” vs. “collision”), symbol substitutions (e.g., “mg” vs. “weight”), or non-critical numerical approximations exist. \\
        \textbullet\ Mark \textbf{False} if key elements are missing/added, process types contradict, start/end states differ significantly, or causal logic is reversed. \\[1em]
        {\textbf{\textcolor{MidnightBlue}{Evaluation Aspects:}}} \\[0.3em]
\begin{tabularx}{\textwidth}{@{}l X@{}}
\textbullet\ \textbf{Force \& Energy} & Includes all relevant forces (type, magnitude, direction), balance conditions, and energy transformations. \\
\textbullet\ \textbf{Process Detail} & Applies to dynamic cases: process\_name, trigger, start\_condition, end\_condition, process\_description. \\
\textbullet\ \textbf{Entity State} & Match of entity/situation structure: consistent id, situation\_id, and state\_type (equilibrium vs. dynamic). \\
\textbullet\ \textbf{Process Link} & Compare related\_processes and relation\_type (e.g., triggered, sequential). \\
\textbullet\ \textbf{State Change} & Match initial and final states: initial\_physical\_state, final\_physical\_state (position, velocity, acceleration, energy). \\
\end{tabularx} \\[0.5em]

        {\textbf{\textcolor{MidnightBlue}{Input Format:}}} \\
        \textbullet\ \textbf{Ground Truth}: \{ground\_truth\} \\
        \textbullet\ \textbf{Large Language Model}: \{large\_language\_model\_result\} \\[0.5em]

        {\textbf{\textcolor{MidnightBlue}{Output Format (JSON Template):}}}
\begin{verbatim}
{
  "force_and_energy": <boolean>,        // e.g., True
  "process_detail": <boolean>,          // e.g., False
  "entity_state": <boolean>,            // e.g., True
  "process_link": <boolean>,            // e.g., False
  "state_change": <boolean>             // e.g., True
}
\end{verbatim}
    \end{tcolorbox}
    \caption{Evaluation prompt used for judging alignment between MLLM-predicted Process Formulation result and ground truth across key physical factors.}
    \label{fig:process_judgement_prompt}
\end{figure*}

\begin{figure*}[ht!]
    \centering
    \begin{tcolorbox}[
        colback=white,
        colframe=black,
        enhanced jigsaw,
        listing only,
        listing options={basicstyle=\rmfamily},
        title={\large Judgement Prompt for Solution Derivation}
    ]
        {\textbf{\textcolor{MidnightBlue}{Instructions:}}} \\
        Please evaluate whether the answer generated by the large language model aligns with the provided ground truth. \\[0.5em]

        {\textbf{\textcolor{MidnightBlue}{Evaluation Criteria:}}} \\
        \textbullet\ Mark \textbf{True} if the generated answer is essentially consistent with the ground truth. \\
        \textbullet\ Mark \textbf{True} if minor differences in formatting or phrasing while the two answers are logically equivalent. \\
        \textbullet\ Mark \textbf{False} if the generated answer deviates in a way that changes its meaning or correctness. \\[0.5em]
        
        {\textbf{\textcolor{MidnightBlue}{Input:}}} \\
        \textbullet\ \textbf{Ground Truth}: \{ground\_truth\} \\
        \textbullet\ \textbf{Large Language Model}: \{large\_language\_model\_result\} \\[0.5em]

        {\textbf{\textcolor{MidnightBlue}{Output Format:}}}
\begin{verbatim}
<boolean>                           // e.g. True
\end{verbatim}
    \end{tcolorbox}
    \caption{Evaluation prompt used for judging alignment between MLLM-predicted Solution Derivation result and ground truth.}
    \label{fig:answer_judgement_prompt}
\end{figure*}

\section{Case Study}
\label{app:case_study}
\begin{figure*}[htb!]
  \centering
  \includegraphics[width=\linewidth]{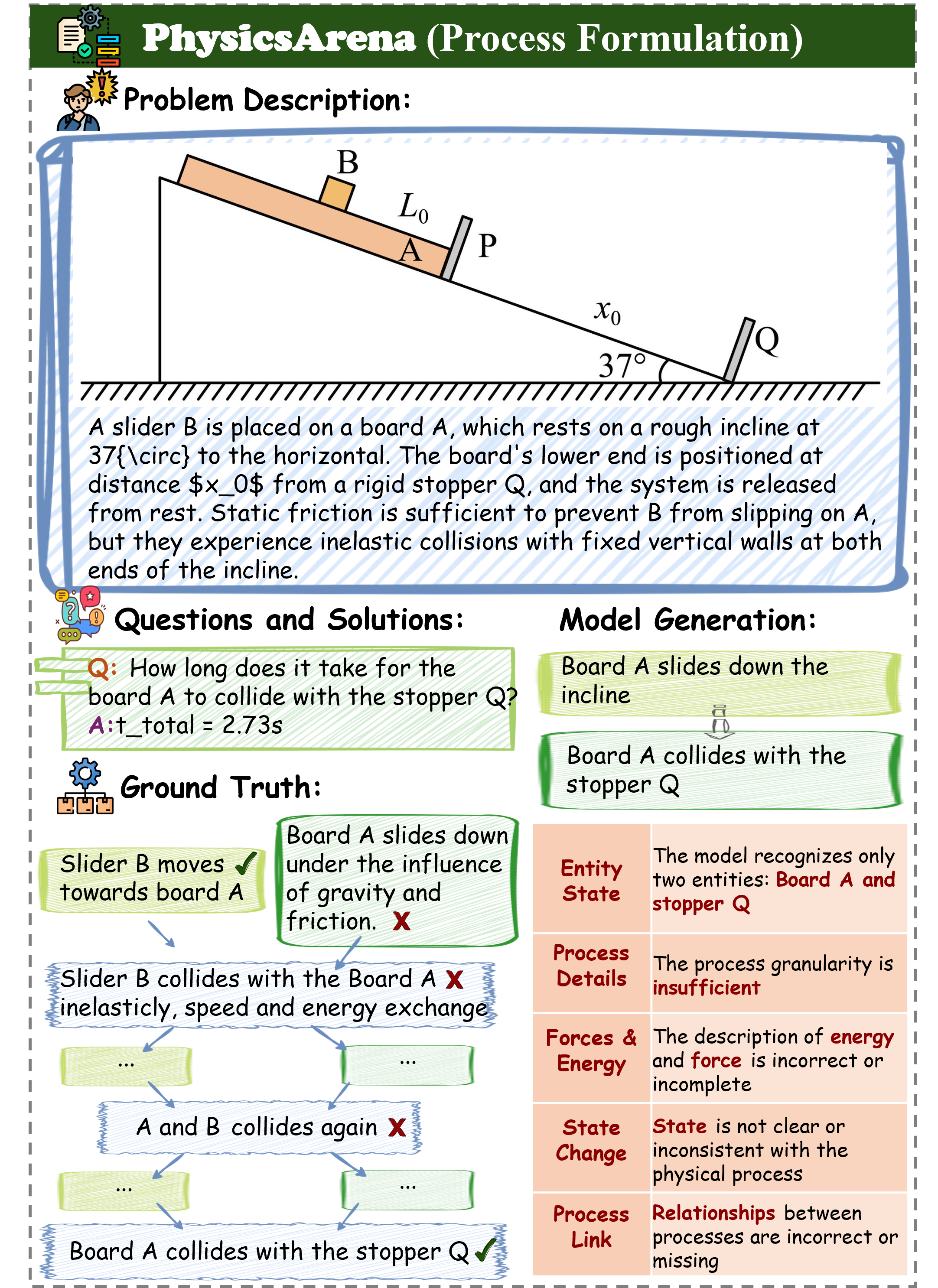}
  \caption{A representative bad case of Process Formulation. Full problem see Figure~\ref{fig: problem_example_2}.}
  \label{fig:bad-case-process-1}
  \vspace{-4mm}
\end{figure*}

In addition to the case study on variable identification presented and analyzed in the main text (Figure~\ref{fig:bad-case-variable-1}), we also provide an example of Process Formulation in Figure~\ref{fig:bad-case-process-1}, which corresponds to the analysis discussed in the main text.

\section{Correlation Analysis}
\label{app:correlation_analysis}
\begin{figure*}[htbp]
  \centering
  \subcaptionbox{Easy\label{subfig:var_easy}}[0.32\linewidth]{%
    \includegraphics[width=\linewidth]{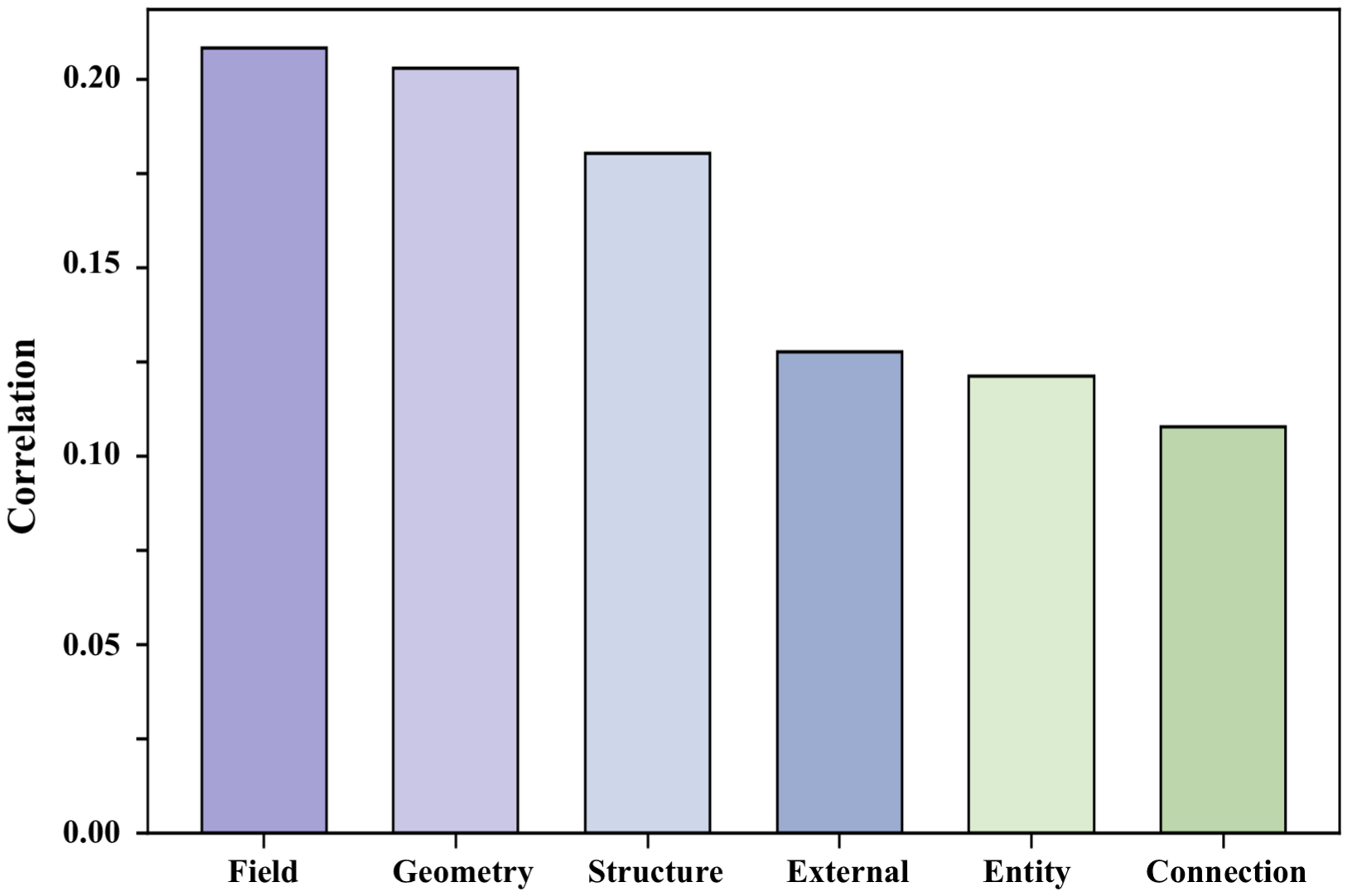}}%
  \hfill
  \subcaptionbox{Medium\label{subfig:var_medium}}[0.32\linewidth]{%
    \includegraphics[width=\linewidth]{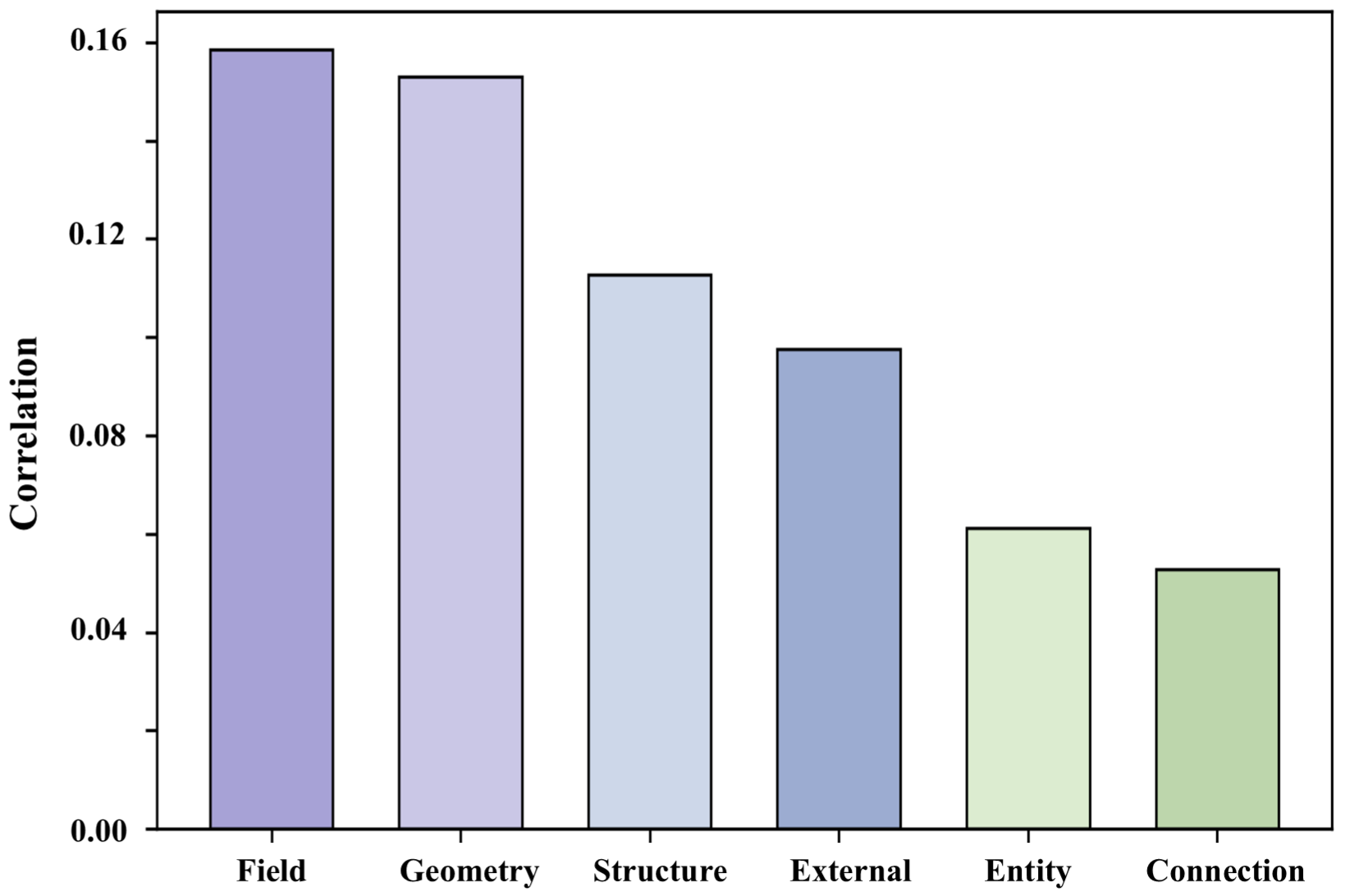}}%
  \hfill
  \subcaptionbox{Hard\label{subfig:var_hard}}[0.32\linewidth]{%
    \includegraphics[width=\linewidth]{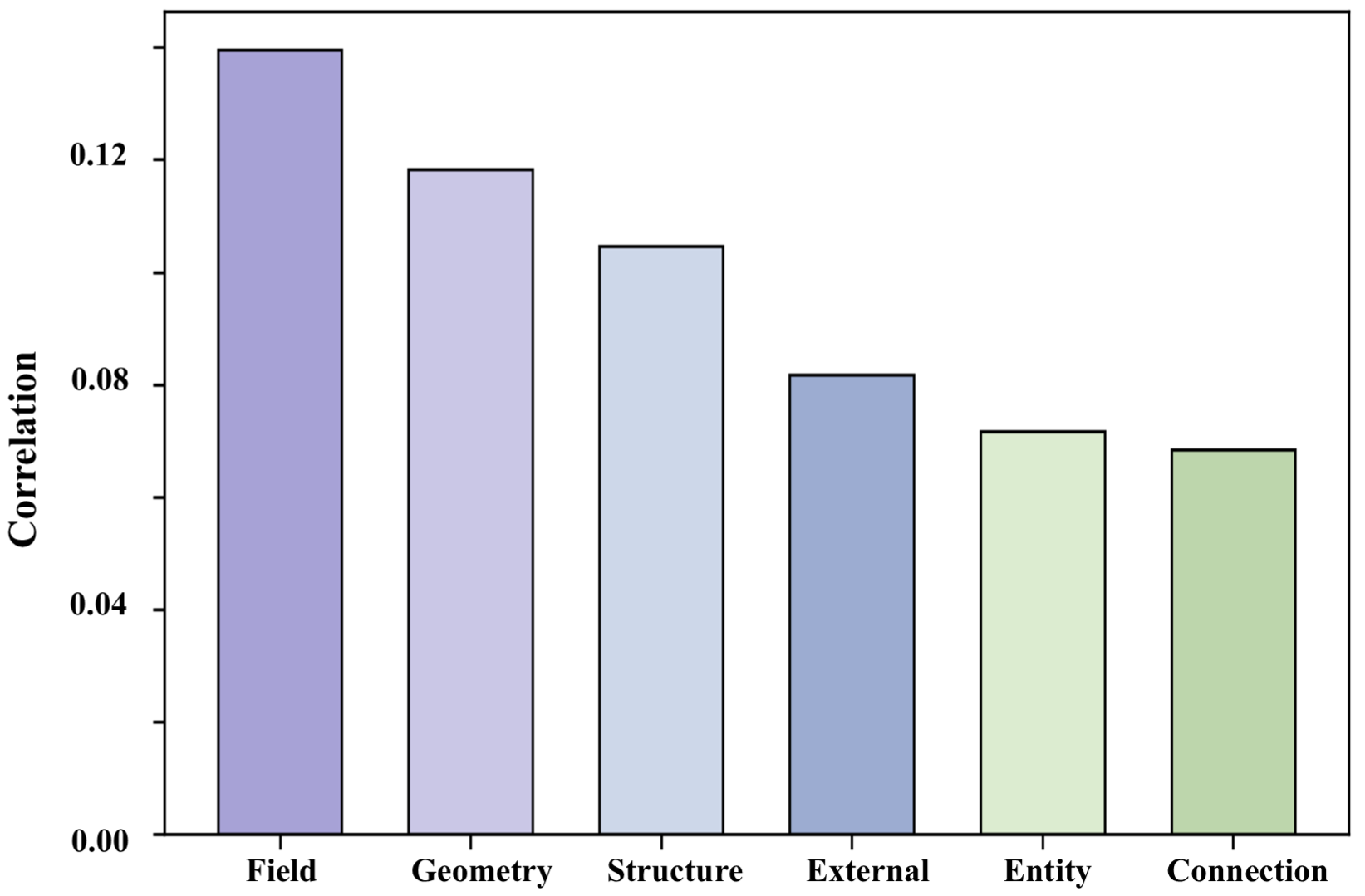}}

  \vspace{-4mm}

  \subcaptionbox{Easy\label{subfig:proc_easy}}[0.32\linewidth]{%
    \includegraphics[width=\linewidth]{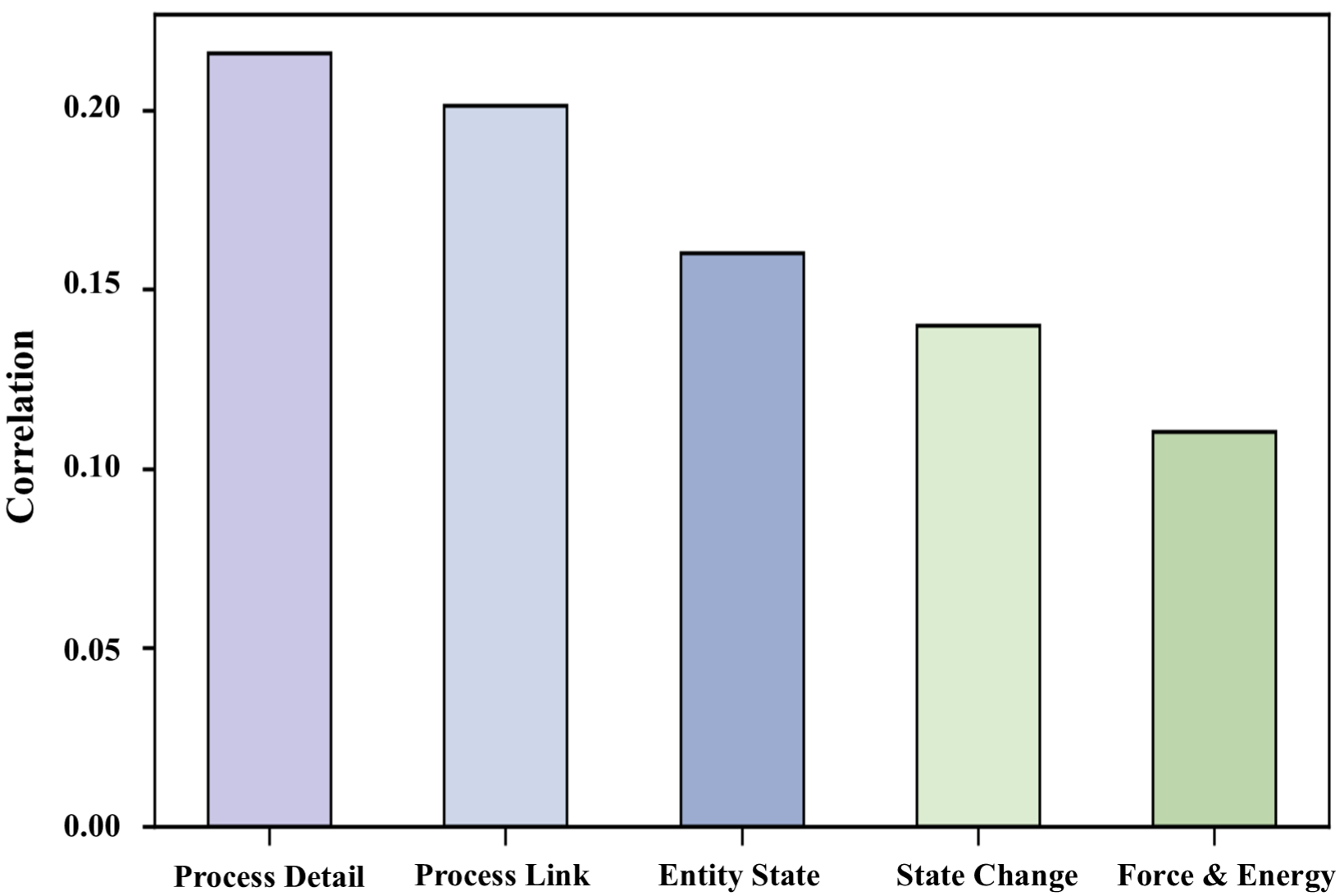}}%
  \hfill
  \subcaptionbox{Medium\label{subfig:proc_medium}}[0.32\linewidth]{%
    \includegraphics[width=\linewidth]{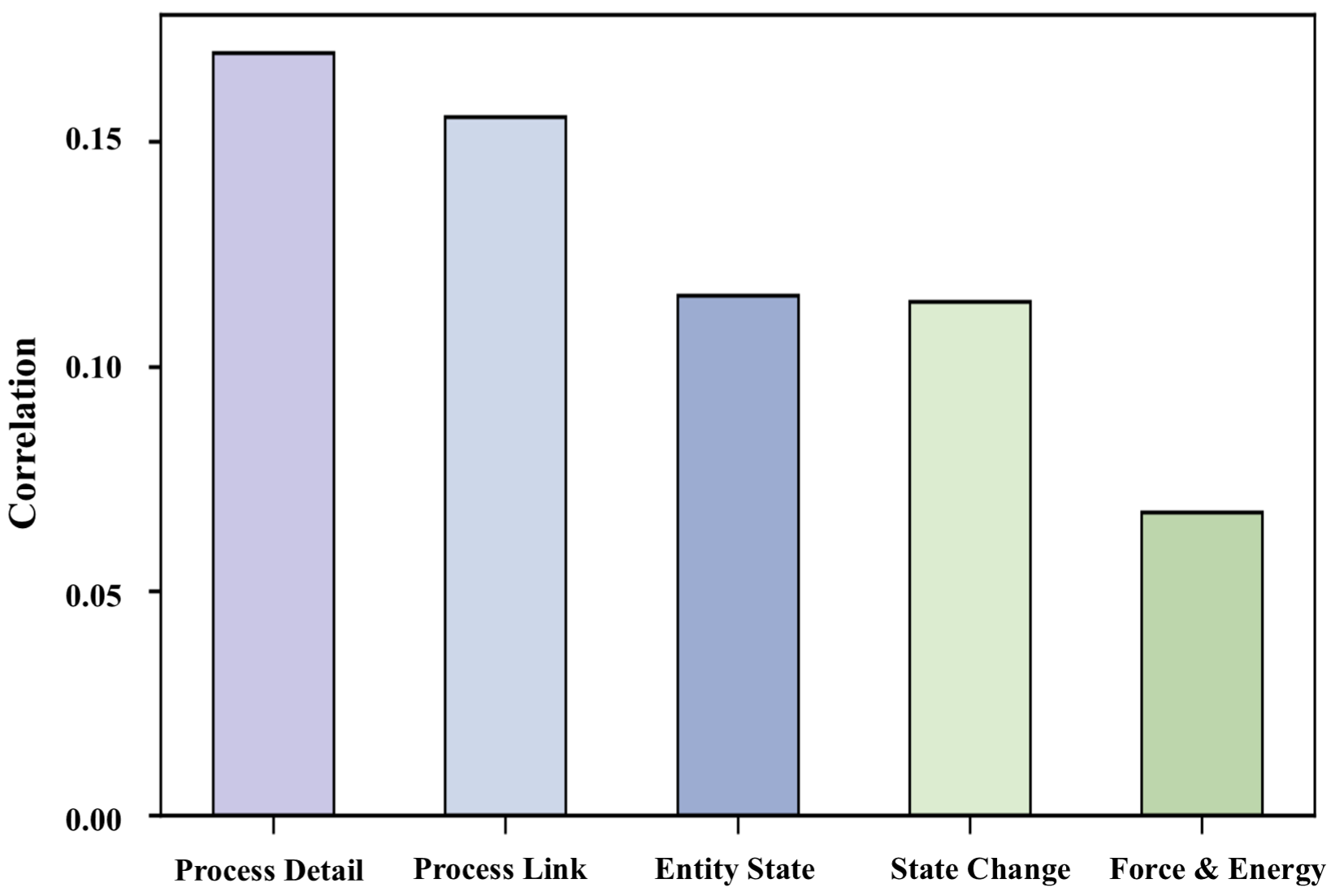}}%
  \hfill
  \subcaptionbox{Hard\label{subfig:proc_hard}}[0.32\linewidth]{%
    \includegraphics[width=\linewidth]{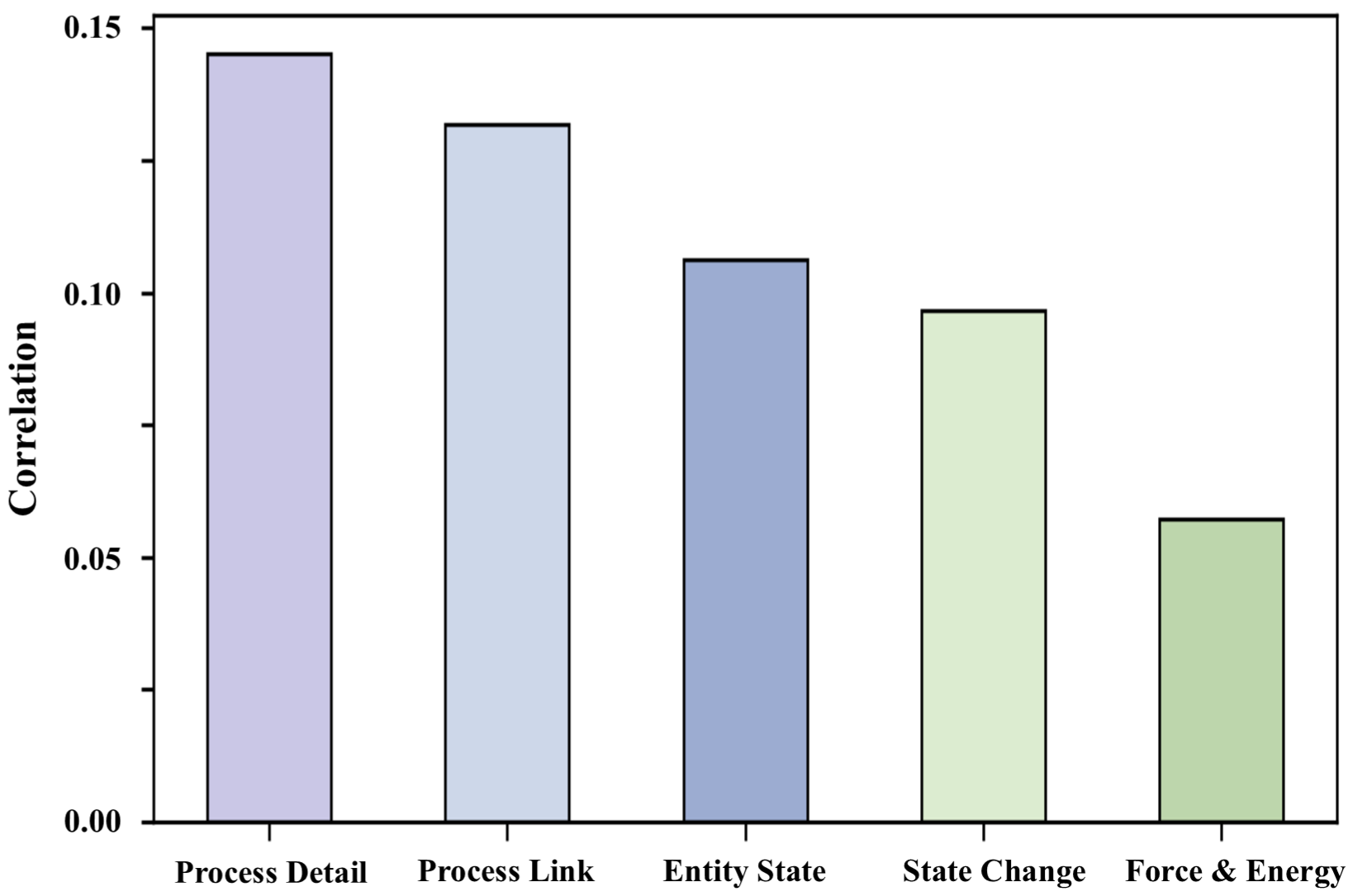}}

  \caption{Correlations between two categories of cognitive factors and solution accuracy across difficulty levels.  
           (a–c) Variable-Identification factors; (d–f) Process-Formulation factors.}
  \label{fig:corr_factors_difficulty}
\end{figure*}
We analysed, separately for easy, medium and hard problems, (1) the correlation between variable-identification factors and solution accuracy and (2) the correlation between process-formulation factors and solution accuracy(Figure~\ref{fig:corr_factors_difficulty}). In every case the rank order of the correlations was preserved, indicating that each factor’s relationship with final accuracy is highly robust.

\end{document}